\documentclass[10pt,twocolumn,letterpaper]{article}

\usepackage{iccv}
\usepackage{times}
\usepackage{epsfig}
\usepackage{graphicx}
\usepackage{amsmath}
\usepackage{amssymb}
\usepackage{subcaption}
\usepackage{enumitem}

\usepackage{booktabs}
\usepackage{mathtools}
\usepackage{algorithm}
\usepackage{algpseudocode}
\usepackage{bm}

\usepackage{todonotes}
\setuptodonotes{size=tiny, backgroundcolor=orange!20}

\DeclareMathOperator*{\argmax}{arg\,max}

\usepackage[labelfont=bf]{caption}

\usepackage[pagebackref=true,breaklinks=true,letterpaper=true,colorlinks,bookmarks=false]{hyperref}

\usepackage[capitalize]{cleveref}
\crefname{section}{Sec.}{Secs.}
\Crefname{section}{Section}{Sections}
\Crefname{table}{Table}{Tables}
\crefname{table}{Tab.}{Tabs.}

\usepackage[capitalize]{cleveref}
\crefname{section}{Sec.}{Secs.}
\Crefname{section}{Section}{Sections}
\Crefname{table}{Table}{Tables}
\crefname{table}{Tab.}{Tabs.}

\usepackage{adjustbox}
\usepackage{array}

\newcolumntype{R}[2]{%
    >{\adjustbox{angle=#1,lap=\width-(#2)}\bgroup}%
    l%
    <{\egroup}%
}

\iccvfinalcopy 

\title{Thinking Like an Annotator: Generation of Dataset Labeling Instructions}
\author{Nadine Chang\\
Carnegie Mellon University\\
{\tt\small nadinec@cmu.edu}
\and
Francesco Ferroni\\
Nvidia\\
{\tt\small francescoferroni1@gmail.com}
\and
Michael J. Tarr\\
Carnegie Mellon University\\
{\tt\small michaeltarr@cmu.edu}
\and Martial Hebert \\
Carnegie Mellon University \\
{\tt\small mhebert@andrew.cmu.edu}
\and Deva Ramanan\\
Carnegie Mellon University\\
{\tt\small deva@andrew.cmu.edu}
}
\begin{document}

\twocolumn[{%
\renewcommand\twocolumn[1][]{#1}%
\maketitle

\begin{center}
\vspace{-0.4cm}
    \includegraphics[width=\linewidth,keepaspectratio=true]{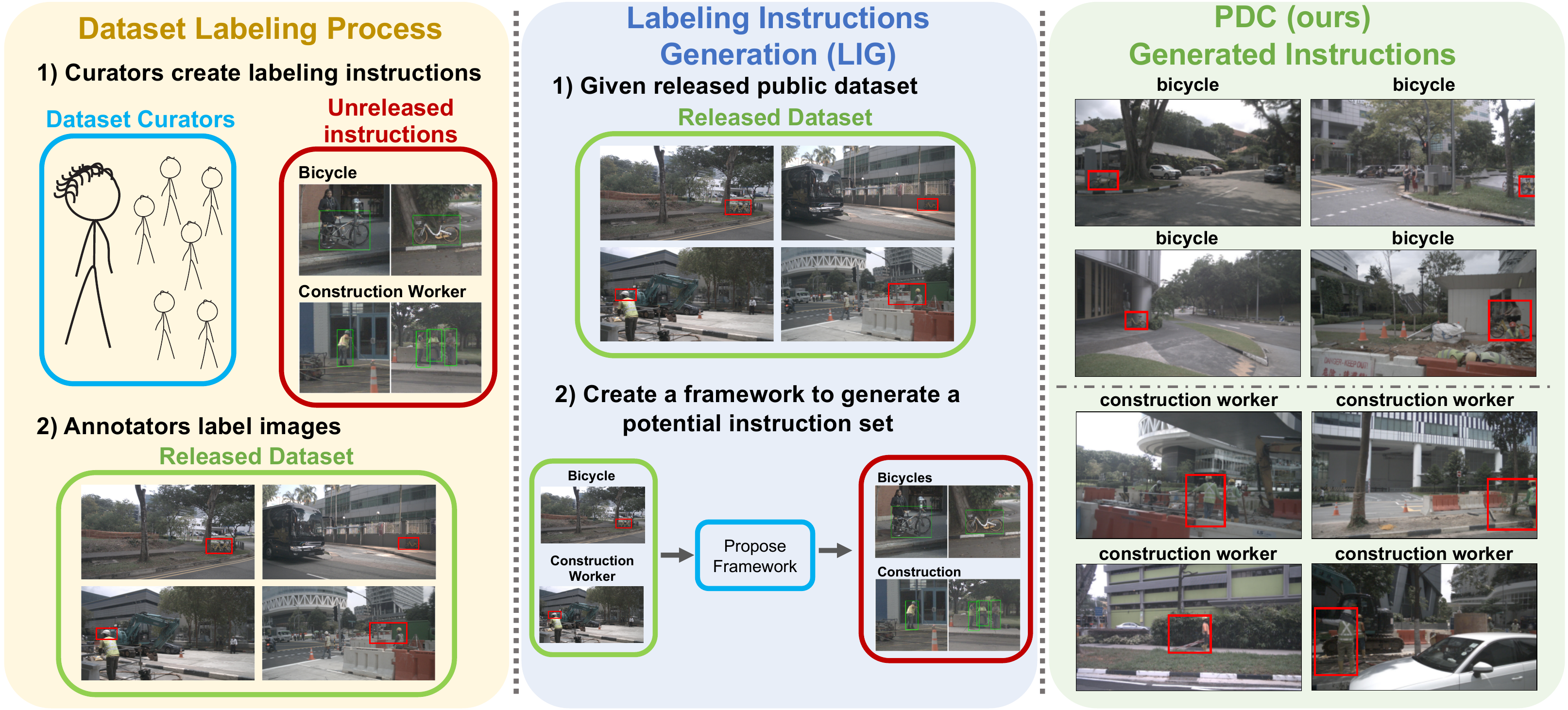}
    \captionof{figure}{Left: The manual dataset labeling process that creates instructions for annotators is tediously long. Center: we propose LIG to address the lack of public labeling instructions for most datasets. LIG's main objective is to generate instructions given a released dataset. Right: Proxy Dataset Curator (PDC) addresses LIG. We show generated instruction pairs sets here. Note that each image is accompanied by a text phrase to compose an instruction pair.\label{teaser}}
\end{center}

}]


\begin{abstract}
Large-scale datasets are essential to modern day deep learning. Advocates argue that understanding these methods requires dataset transparency (e.g. “dataset curation, motivation, composition, collection process, etc…”)~\cite{gebru2021datasheets}. However, almost no one has suggested the release of the detailed definitions and visual category examples provided to annotators – information critical to understanding the structure of the annotations present in each dataset. These labels are at the heart of public datasets, yet few datasets include the instructions that were used to generate them.
We introduce a new task, \textbf{Labeling Instruction Generation}, to address missing publicly available labeling instructions. In Labeling Instruction Generation, we take a reasonably annotated dataset and: 1) generate a set of examples that are visually representative of each category in the dataset; 2) provide a text label that corresponds to each of the examples. We introduce a framework that requires no model training to solve this task and includes a newly created rapid retrieval system that leverages a large, pre-trained vision and language model. This framework acts as a proxy to human annotators that can help to both generate a final labeling instruction set and evaluate its quality. Our framework generates multiple diverse visual and text representations of dataset categories. The optimized instruction set outperforms our strongest baseline across 5 folds by 7.06 mAP for NuImages and 12.9 mAP for COCO.
\vspace{-0.5cm}
\end{abstract}

\section{Introduction}
\vspace{-0.2cm}
Large-scale datasets are the foundations for almost all modern computer vision tasks. As a field, we often evaluate these datasets based on their released data and annotation quality. Underlying annotation quality is an intensive curation process that is reflected in the labeling instructions (LIs) required to annotate that data.
LIs are typically the result of numerous painstaking discussions aimed at clarifying desired class memberships and at aligning annotations between data curators and annotators. Despite incurring significant time and financial costs, end-state LIs are rarely available. We attempted, but failed, to obtain the LIs for benchmark datasets such as COCO~\cite{lin2014microsoft}, ADE20K~\cite{zhou2017scene}, OpenImages~\cite{krasin2017openimages}. This lack of availability illustrates how our community has overlooked the clear and comprehensive reporting of dataset annotation. Even the influential work \textit{Datasheets for Datasets}~\cite{gebru2021datasheets}, which advocates for dataset transparency, falls short of arguing for the release of LIs. The inaccessibility of instruction policies forms a major gap in efforts towards full transparency and reproduciblity. 

{\bf Why are instructions important and what are their applications?} Beyond providing rich details about classes and the boundaries between classes, accessible LIs helps the research community in multiple ways:

\textit{Reproduciblity.} LIs are critical for understanding foundational concepts such as model generalization and overfitting. Recent efforts in analyzing overfitting on ImageNet~\cite{deng2009imagenet} recreated a held-out test set by reusing the original annotation instructions~\cite{recht2019imagenet}. Such continual validation set generation is impossible without LIs. Indeed, in a real-world application related to autonomous vehicles, a version of our framework was used for ``continual dataset construction''. That is, although LIs were available, generated LIs were used to refine annotation policy during the natural annotator-curator conversation that occurred as more data (and edge cases) were collected and labeled.

\textit{Clarifications.} Analyses of errors in public datasets~\cite{kangfinding} reveal that many `errors' are due to LIs. For example, in one dataset, annotators were instructed to \textit{not label} vehicles that appeared outside of the drivable regions in images~\cite{kang2022}.

\textit{Medical Biases.} In medical imaging, framing biases arise from how instructions are presented to specialist labelers \cite{Itri2018}. Recovering LIs is central to revealing and understanding these biases in medical datasets where LIs are protected, proprietary or private, yet their application has serious real-world consequences.

\textit{Human Studies.} Data collection (protocol) transparency is longstanding in the behavioral sciences because behavior and decisions vary with demand characteristics (instructions) and replication is impossible without protocol information. Publication in these fields requires rigorous documentation of data collection protocols including instructions~\cite{Wilkinson2016}. Building on recent calls for transparency \cite{gebru2021datasheets}, similar standards are warranted for computer vision.
    
\textit{Policy Initiatives.} The media, public, and lawmakers (e.g., EU privacy laws such as GDPR) are all increasingly concerned with data bias and transparency in AI. Yet there has been little discussion of how LIs may inject bias or how users gain a transparent view of labels when provided with LIs. Identifying the issue and introducing a novel task+method are initial steps towards a healthy discussion on this critical facet of the larger policy debate.

We address the lack of publicly available LIs by proposing a new task, \textbf{Labeling Instruction Generation (LIG)}. 

We first identify the typical composition of the annotation instructions for a given dataset, for example, the set from NuImages~\cite{nuscenes} as shown in~\cref{teaser}. Note that LIs are often {\em multi-modal sets} of text descriptions plus visual examples. Text descriptions provide both a label and detailed definitions of classes and attributes, synonyms, and descriptions of corner cases. Visual examples are typically composed of prototypical representations and rarer class sub-type representations. For instance, a prototypical image for motorbike is a two-wheeled bike and a rarer sub-type is an image of a motorbike with three wheels. Adding to the effectiveness of the instruction set, visual examples are frequently shown from various viewpoints and scales. The combination of both text descriptions and image examples provides a compelling, informative, and generally applicable dataset labeling instruction set. Reflecting this degree of exposition, we focus on generating LIs that are similarly composed of both text descriptions and visual examples. 

Our proposed task, LIG, starts with a given annotated dataset, which at minimum consists of categorical labels and the associated images/bounding boxes that contain the objects referred by these labels. Our objective is to generate a set of category labeling instructions that can be shown to new annotators and effectively demonstrates the desired types of image classes to be labeled in new images. To generate informative instructions, the final instruction set must have, for each category both a set of visual examples that are representative of that category and a set of text labels that corresponds to each generated visual example. The final result is a set of {text, image} pair(s) for each category (\cref{teaser}).

To solve LIG, dataset curators and annotators engage in a painstaking cycle of instruction policy refinement (in this way dataset curators are effectively dataset annotators too). Curators provide an initial instruction set to annotators, who begin to label the dataset. Annotators inevitably run into confusing cases, where they unsure as to how to label a given object. Ambiguous cases are brought back to the curators, who incorporate these new cases into an updated instruction set. This back and forth between curators and labelers continues and results in a detailed instruction policy that may be as expensive to obtain and as valuable as the annotations themselves \textit{given} that policy. In that such instruction policies are rarely made available, our goal is to generate these instruction sets to increase the transparency and utility of public datasets. 


Beyond surfacing instruction policies for datasets in which LIs are not available, given the importance of instruction policies, it is desirable to develop an alternative, non-manual solution to LIG. We propose a computationally efficient method that is a proxy for dataset curation. Large-scale vision and language models (VLMs), such as CLIP, ALIGN, and Florence \cite{radford2021learning,jia2021scaling,yuan2021florence}, provide text and image representations that yield robust results for a variety of tasks in the open-world. We leverage VLMs to build a framework that rapidly traverses a dataset and retrieves the best {text, image} pair(s) that are representative of a given class. Our framework, which requires no back propagation and only inference level modifications, consists of three components: 1) An image representations database constructed from a dataset’s images converted into representations via a pre-trained VLM; 2) An image retrieval system that rapidly queries through this database; 3) Multi-modal prompts that can be used with a pre-trained VLM. Condensing text and image representations into a single query via multi-modal fusion, we show that multi-modal queries are essentially free, without additional compute. 

Our algorithmic framework, named \textbf{Proxy Dataset Curator} or PDC, is intuitively demonstrated in \cref{fig:algo}. PDC is a greedy algorithm that, on a high-level, searches for the best object images and best text description. Images and texts are paired up and used as queries for image retrieval on the entire training dataset. The set of pairs that achieves the best retrieval performance is chosen as the final instruction set. Importantly, the pairs are surfaced from a training set and final mAP evaluation on held-out test set is reported. We perform 5 fold training and testing for evaluation stability. PDC is described in ~\cref{method:code} and fully detailed in pseudocode in ~\cref{code_app} of the Appendix.

\textbf{How do we evaluate labeling instructions?}
The gold-standard evaluation of LIs is in a human annotation setting. The ideal design would include generated LIs for a human annotation task and inspection of the quality of the resulting annotations as collected on a full dataset. Since this evaluation at scale is prohibitively expensive, impractical, and time-consuming, we instead perform two scalable evaluations. First, we perform a human experiment with forced choices between pairs of candidate instruction sets: the original instructions from NuImages versus our generated LIs. Participants choose which of the two sets is preferred for future annotation tasks. Second, we evaluate our generated instructions for both COCO and NuImages on a held-out {\em annotated} test set by building a multi-modal retrieval engine that returns images that are likely to contain the object class of interest. We then report mAP on these retrieved images. These two evaluations reveal that our PDC method is an effective means for LIG. 

For retrieved images from NuImages, PDC generated LIs outperform our strongest baseline by 7.06 mAP and the original NuImages instructions by 12.9 mAP. Similarly, for retrieved images from COCO, PDC generated LIs outperform our strongest baseline by 12.9 mAP (as noted, we tried and failed to obtain the original COCO instructions for comparison).
Second, for NuImages, across all classes, human evaluators preferred our generated instructions over the original instructions 44\% of the time. Importantly, this indicates that our generated instructions are visually almost as good as the original instructions. Thus, while this behavioral experiment demonstrates that participants are nearly as likely to prefer our generated LIs as the originals, our quantitative evaluation demonstrates that our PDC generated LIs provide additional benefits. In particular, our generated LIs outperform both baseline and original LIs (due to better consideration of corner and other boundary cases).

\textbf{Limitations.} While these results are promising, we acknowledge several limitations to our present work. First, our framework focuses on generating text and image pairs because such pairs are the most commonly used in real-world LIs. Thus, although richer multi-modal instructions are a potential future direction, we view text+images as the best first step in addressing this new problem. Second, generated text instructions may sometimes be less nuanced and/or detailed as compared to human generated text describing visual classes. We expect that rapid advances in LLMs and VLMs will enable more expressive generated text instructions in the near future. Third, while our framework does generate corner cases, our current implementation does not include negative examples. This is in large part because negatives are presently difficult to represent in both LLMs and VLMs. We expect that progress in this area will enable us to consider negatives in future versions.

In sum, we view our contributions as follows. First, we highlight LI inaccessibility as an overlooked problem in publicly available datasets that directly impacts transparency, fairness, and reproducibility. Second, we propose a new task, LIG, to address this problem by generating multi-modal instructions (visual examples plus text) for existing datasets that lack LIs due to legal or privacy concerns or simple refusal to publish. Third, we propose PDC, a new framework for solving LIG, that acts as a proxy for curators and annotators in the creation of LIs. We experimentally show that our framework, which requires only model inference time changes, is fast, scalable, and efficient because: 1) we can create a pre-computed database index; 2) we do not require model training; 3) we utilize a fast cosine similarity operation. We establish the effectiveness of PDC through both computational and human experiments. 

\section{Related Works}
\vspace{-.1cm}
{\bf Dataset Instructions and Transparency}
Most influential datasets do not include the annotation instructions that the dataset curators provided to the annotators. COCO, a key detection dataset, crowd sourced annotations using Amazon Mechanical Turkers~\cite{lin2014microsoft}. As such, the COCO curators had to provide instructions. However, as mentioned, we were unable to obtain the unreleased instructions for COCO. ImageNet challenge~\cite{russakovsky2015best}, a key classification dataset, collected candidate images by querying the internet and uses ``both automatic and manual strategies" to clean up search results. None of the manual steps are publicly released. Even today, datasets such as OpenImages~\cite{openimages}, need annotators to verify labels. Again, no instructions passed on to annotators are publicly available. In contrast, many datasets completely avoid generating annotation instructions. TinyImages uses synsets from WordNet~\cite{beckwith2021wordnet} and scraped the web to collect 80 million images. However, results are not manually verified, and recently the dataset was publicly removed due to heretofore unflagged inappropriate content~\cite{birhane2021large}. The lack of transparency with such high-profile datasets has motivated advocates to compile a list of objectives that would facilitate transparency~\cite{gebru2021datasheets}. This list focuses on expanding details in dataset composition and the collection process that should be answered by dataset curators; however, it omits the process of creating labeling instructions or subsequent public release. The instructions provided to dataset annotators are valuable and expensive to compile and yet completely overlooked and unreleased.   

{\bf Improving Labeling Instructions for Better Annotations}
One approach to improving annotation quality is to learn whether to prompt an annotator for category labels or object attribute labels~\cite{kovashka2011actively}. Similarly, one can learn what types of annotations (e.g., bounding boxes vs. tight or loose segmentations) are sufficient for learning a category~\cite{jain2013predicting}. An alternative approach is to examine an annotator's visual interpretations of object attributes described through text~\cite{kovashka2015discovering}. Finally, LVIS completely bypasses creating labeling instructions and instead ask annotators to `point' to an object~\cite{bearman2016s} and label that object's self-defined category~\cite{gupta2019lvis}. 
The large body of work that aims to analyze and improve the annotation process illustrates the significance of high-quality annotation instructions and pipelines. Despite this importance, rarely do datasets publicly release their annotation code, instructions, or pipeline. 

{\bf Large-Scale Vision and Language Models} 
Recent large-scale Vision and Language models (VLMs) trained on extremely large amounts of data align images and text into a common representation space~\cite{radford2021learning,jia2021scaling,li2022grounded,yuan2021florence}. This alignment allows for simple similarity search between two modalities (text, image). Thus, VLMs demonstrate remarkable general domain learning and zero-shot capabilities. Several works have built on top of VLMs, applied them successfully on traditional tasks, and seen clear improvements~\cite{skantze2022collie,zhou2021learning}. One work leverages a VLM to learn personalized concepts from users (e.g. `my favorite skirt') and requires a user to input a specialized instruction set~\cite{cohen2022my}. Note that our task is vastly different, as we aim to generate categorical instruction sets automatically. From these successes, we observe that these models are increasingly robust and difficult to outperform. Interestingly, works show that finetuning on VLMs tend to hurt its generalization~\cite{wortsman2022robust,kumar2022fine}.

{\bf Multi-modal Training}
Multi-modal training with both images and language extends VLMs. Captioning tasks, where an image is provided and a caption is asked for, jointly train with both text and language~\cite{antol2015vqa,goyal2017making,marino2019ok}. One multi-task model trains on several vision and language datasets for VQA and captioning tasks~\cite{lu202012}. Scene graphs, which are structured graphs depicting relationships between attributes and objects, also includes multi-modal text and image training~\cite{johnson2015image,yang2018graph,xu2017scene}. Scene graphs have been used for image retrieval as well~\cite{johnson2015image}, but require training and creation of a complex graph structure. Finally, a new retrieval task relies on an input image and a modification text that explains how to change the input image~\cite{vo2019composing,liu2021image}. However, these models require extensive training and image to text pairs. Moreover, these models do not attempt to align both image and text representations into a single space. 


\vspace{-0.2cm}
\section{Method}
\vspace{-0.2cm}

We split our method section into three parts: 1) The creation of our database index that allows us to do rapid image retrieval, 2) the various query policies and functions that can be performed on our database index, 3) our proposed algorithmic framework that utilizes the index and selected query policies to generate our final labeling instruction pairs.
\vspace{-0.1cm}



\subsection{Creating an Index}
\begin{figure}[h!]
\centering
\includegraphics[width=.9\linewidth]{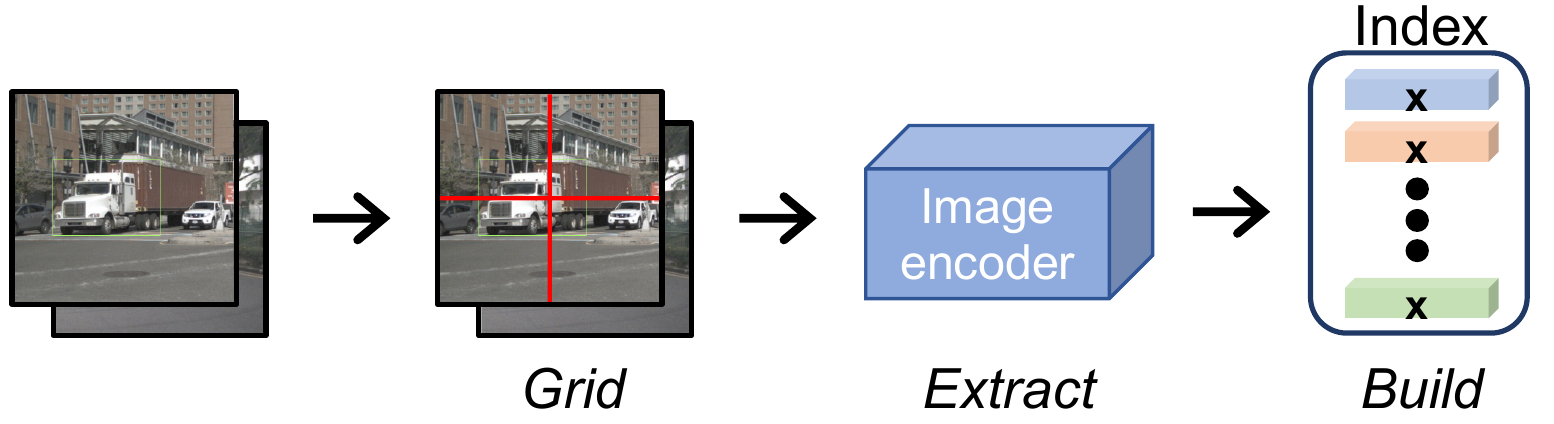}
\caption{An overview of building our database index.}
\label{fig:overview} 
\vspace{-0.2cm}
\end{figure}

\vspace{-0.2cm}
\label{sec31} 
First, we acknowledge that building a queryable database index with VLM embeddings is a rather simple and intuitive setup, but have no knowledge of any works that have built such an index. Thus, we detail our index building below. We build a database index that contains the visual embeddings corresponding to the annotated dataset we would like to generate labeling instructions for. We create both testing and training indexes, which will not be altered once created. To extract all embeddings, we utilize a pre-trained image encoder from a VLM. These embeddings are trained using cross-modal contrastive losses, such that cosine similarities result in meaningful cross-modal similarity. Our index building pipeline can be seen in \cref{fig:overview}. 

{\bf Grid and Extract.}
One can create an index of visual embeddings by extracting them from whole images. However, to focus on objects of varying scale (e.g. small and large objects), we additionally grid each image into patches. For further diversity, we use odd numbered grid sizes where $g=1,3,5,7,9$ so that patches do not overlap significantly. Each image is cropped into $g^2$ number of equal patches for a final $M$ total patches per image $i$, where $M = \sum g^2$. Patches are passed into a visual encoder to obtain visual embeddings. Each image $i$ contains $M = 165$ embeddings. 

{\bf Build Index.}
Extracting all embeddings yields $N\times (M+1)$ embeddings, which increases our database by $165$ times its original size. Loading all embeddings into run-time memory is infeasible and unscalable. We address this issue by using the FAISS library~\cite{johnson2019billion}. We create a FAISS index which enables fast searches through an on-disk memory stored index. For example, a single search takes approximately $50ms$ on NuImages.
\vspace{-0.1cm}

\subsection{Query Policies}
\label{sec32}
\vspace{-0.2cm}
Here, we describe the various query policies that can be used to query our database index. When we query the index, we get a list of relevant images back from the index that is ranked from most similar to least similar to the query.

\begin{figure}[ht!]
    \centering
    \includegraphics[width=\linewidth]{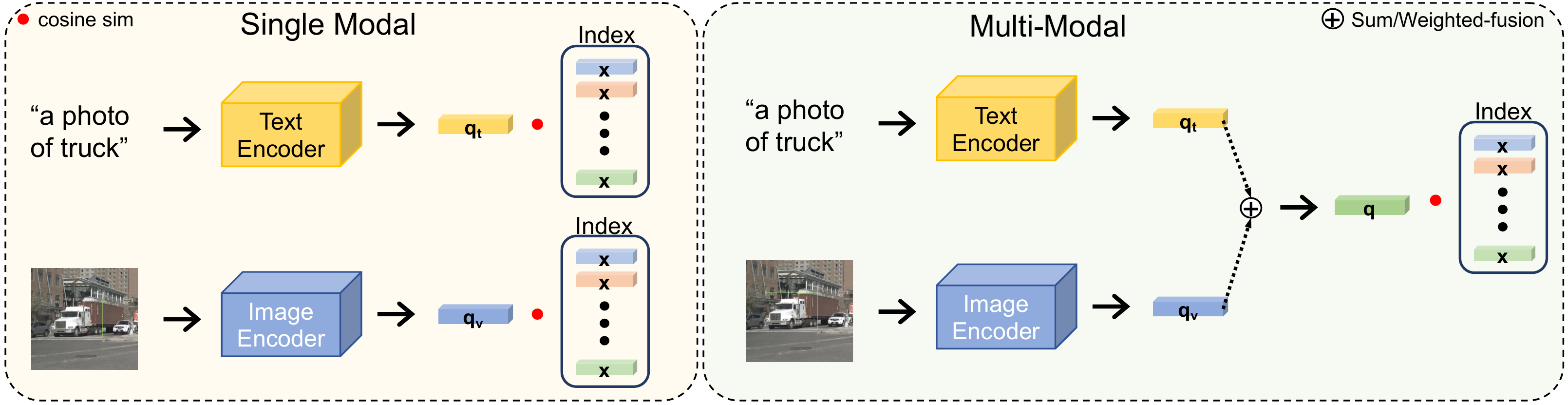}
    \caption{Left: A single modal query with text or visual queries. Right: A multi-modal query that combines text + visual queries into a single query.}
    \label{fig:modal}
    \end{figure}

\textbf{Single Modal.} First, consider queries from a single modality $m$ (text or visual). Let $q_m \in R^D$ be the corresponding VLM embedding. We compute its similarity with a visual embedding $x$. The simplest method is to compute the cosine similarity between a query from modality $m$ (e.g. text or visual) and an example $x$. We define the method as $SingleScore$. An example is seen in ~\cref{fig:modal}.
\begin{align}
  \mathrm{SingleScore}(q_m, x)=\text{cos}(q_m,x) &= {\hat q}_m^T {\hat x}, \\
  \text{where} \quad  {\hat q} = \frac{q}{\sqrt{q^Tq}}, {\hat x} = \frac{x}{\sqrt{x^Tx}}.
\end{align}

\textbf{Multi-Modal.} Second, consider the case with queries from two modalities, text and visual. Let $q_t \in R^D$ and $q_v \in R^D$ be the corresponding text and visual embeddings from a multi-modal model. We explore all possible ways to perform multi-modal query (\cref{fig:modal}).

{\bf Early-fusion: Sum.} We consider an early fusion method~\cite{clinchant2011semantic}, where two modalities are fused. In practice, this means we compute a single embedding. We show that this is also equivalent to late fusion, where we combine results of two modalities' searches. However, we present our method as an early fusion method because it is computationally cheaper - computing one set of score instead of two.
\begin{align}
  \text{SumFusion}(q_t,q_v,x) &= \text{cos}(q,x), \quad  q = {\hat q}_v + {\hat q}_t\\
                                &= {\hat q}_v^T {\hat x} + {\hat q}_t^T {\hat x} \\
  &= \text{cos}(q_v,x) + \text{cos}(q_t,x).
\end{align}
{\bf Early-fusion: Weighted.} SumFusion gives equal weight to both text and visual queries. However, consider the case where we weigh one modality differently by using a weighted average, where weight $w = \text{cos}(q_t,q_v)$. This weighs each text/visual query by how close one is to the other. When $w=1$, this indicates $q_t$ and $q_v$ are the same and combining text and visual queries is not necessary. Thus our weighted query assigns $q=q_t$. When $w=0$, this indicates $q_t$ and $q_v$ are perpendicular. That is, they are not the same nor are they the opposite, but the text and visual embeddings are in between. Thus, our weighted query defaults to averaging, which is equivalent to SumFusion. \begin{align}
  \text{WF}(q_t,q_v,x) &= (1-w)\text{cos}(q_v,x) 
  + (1+w)\text{cos}(q_t,x)\nonumber\\ 
   &= \text{cos}(q,x) \\
   &\quad \text{where} \quad q = (1-w){\hat q}_v + (1+w){\hat q}_t \nonumber
\end{align}

\vspace{-.2cm}

{\bf Late-fusion: Inverse Rank.} Empirically, we find that cosine similarities computed by textual queries are consistently higher than those from visual queries. This suggests that naive combinations such as max or averaging scores give too much weight to text queries. As an alternative, we explore another late-fusion method~\cite{clinchant2011semantic}, where results, not queries, across modalities are combined. Our proposed late-fusion method adopts an approach similar to mean reciprocal rank (MRR) metrics used to summarize retrieval across multiple queries~\cite{Craswell2009}. 
Intuitively, such metrics measure the (harmonic) average rank of correctly retrieved examples. 
We rank example $x$ according to different queries and compute the (inverse) harmonic mean of individual query ranks: $$\text{RankFusion}(q_t,q_v,x) = \frac{1}{\text{Rank}(q_v,x)} + \frac{1}{\text{Rank}(q_t,x)},$$
where $\text{Rank}(q,x) \in \{1,2,\ldots\}$ is the position of example $x$ in the ranked list of retrieved examples given query $q$.

{\bf Late-fusion: Naive.} A less complex version of RankFusion is to iteratively take one result from the ranked returns of text and visual queries until we have obtained the total number of desired returns.

We note that late-fusion methods scale linearly with the number of queries; fusing 10 queries takes 10x more compute. In contrast, early-fusion methods scale by a constant factor by creating a {\em single} multi-modal query embedding $q$.

\subsection{Combining Multiple Query Results}
\label{method:combine}
\vspace{-0.2cm}
Suppose we have two queries from the same modality (e.g. two $q_t$ or $q_v$) or two queries resulting from the same multi-modal fusion technique (e.g. two $q = \mathrm{fusion}(q_v,q_t$). We fuse the unique results of the two queries, $q_{m0}$ and $q_{m1}$, by simply combining results and reranking. In cases where there are duplicate retrieved items from the two queries returns, we address it with the two potential methods below.  

{\bf Max-fusion.} We max the resulting similarity scores across duplicate items. 
$\text{MaxFusion}(q_{m0},q_{m1},x) = \max(\text{cos}(q_{m0},x),\text{cos}(q_{m1},x))$.

{\bf Avg-fusion.} We average the resulting similarity scores across duplicate items.
$\text{AvgFusion}(q_{m0},q_{m1},x) = \mathrm{avg}(\text{cos}(q_{m0},x),\text{cos}(q_{m1},x))$

\subsection{Combining Patch Scores}
\vspace{-0.2cm}
Because we wish to find small objects, our index is composed of both whole image and patch embeddings. To generate one retrieval score per image, we perform a max across similarities for all patches $X_i$ from image $i$. 
$$\mathrm{PatchFusion}(x, i) = s^i_m := \max_{x \in \text{X}_i} \cos(q_m, x)$$

\renewcommand{\algorithmicrequire}{\textbf{Input:}}  
\renewcommand{\algorithmicensure}{\textbf{Output:}} 

\subsection{PDC Algorithm}
\label{method:code}
\vspace{-0.2cm}

Intuitively, PDC searches through an annotated training dataset and greedily grows a set of (text, image) pairs so as to maximize multimodal retrieval performance on the training set. We present a detailed psuedocode in ~\cref{code_app} of the Appendix, but provide an overview in ~\cref{fig:algo}.


\begin{figure}[ht!]
\centering
\caption{After pre-processing, PDC greedily grows the instruction set for a class until no new pairs improve retrieval performance (or a max limit on the instruction size is met).}
\vspace{-0.2cm}
\includegraphics[width=\linewidth]{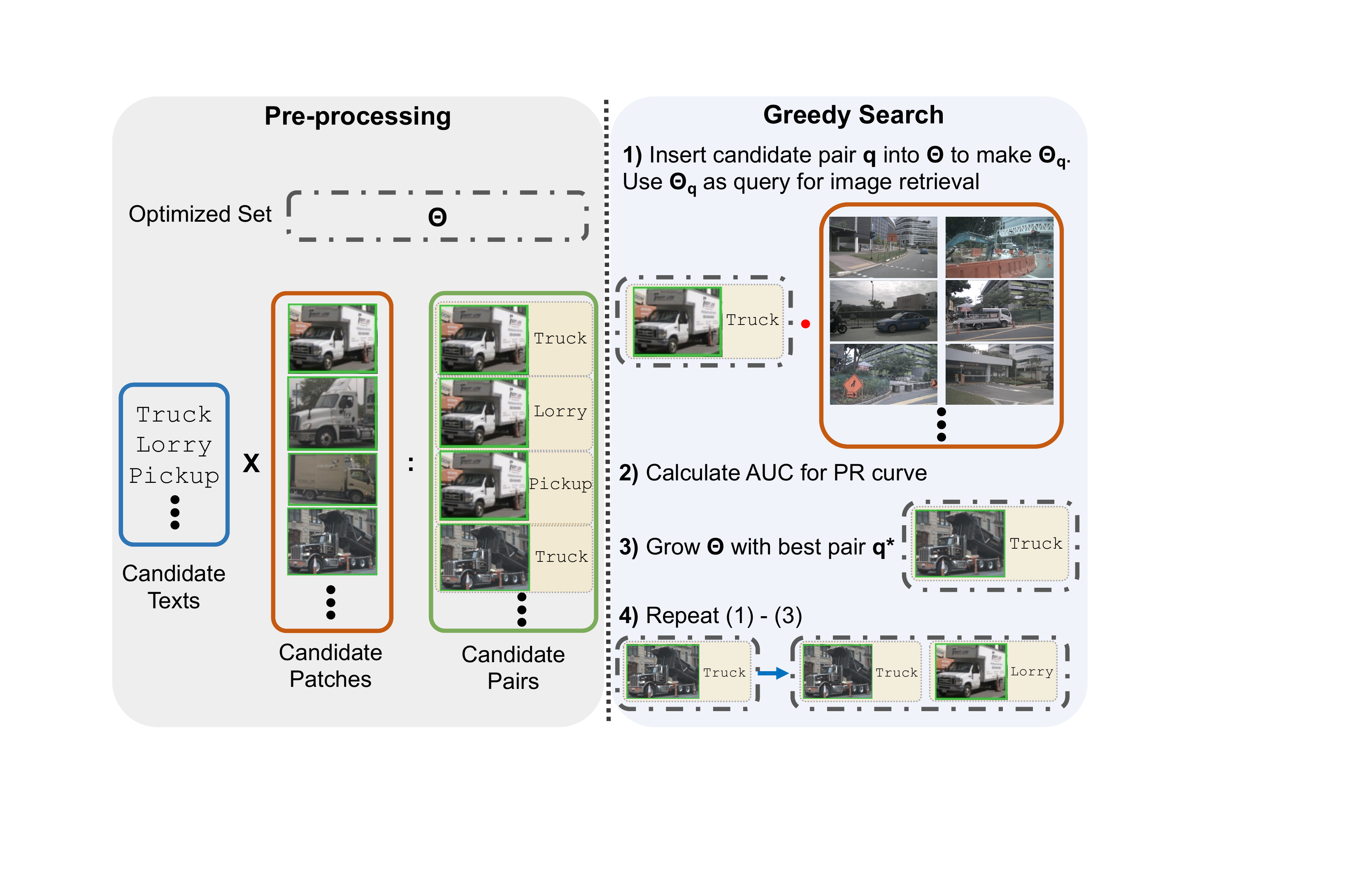}
\label{fig:algo}
\vspace{-0.4cm}
\end{figure}
{\bf Pre-processing.}
Assume there exists a set of descriptive words $L$ associated with a class - e.g., {\tt animal} has the labels/subclasses/synonyms {\tt animal, dog, bird, cat}. Convert each word into a text embedding with our VLM's text encoder $Q_t:=\{f_t(w):w \in L\}$, where $f_t(\cdot)$ is the text encoder. Similarly, assume a dataset of images has been converted into a database of image patch embeddings with our VLM's vision encoder $D:=\{ f_v(p): p \in P_i, i \in I\}$  where $P_i$ is the set of patches extracted from image $i$, and $I$ is the set of training images. We assume our training dataset is {\em annotated} with class bounding boxes, which allows us to define a subset of patches that have been labelled as positive examples; we denote this as $Q_v \subseteq D$. We can now define the set of potential multimodal query pairs as $Q$, the cross product of $Q_t$ and $Q_v$. 

{\bf Setup.} Let $\theta \subseteq Q$ be a set of multimodal (text,image) pairs that represents a candidate labeling instruction (LI) for a particular class. We wish to find a $\theta$ that is ``good", where goodness is measured simply by the image retrieval accuracy when using $\theta$ as a (fused)  multimodal query set. In theory, one could perform an exponential search over all possible subsets of $Q$, evaluate the retrieval accuracy of each candidate LI, and report the subset with the best accuracy. Because this is far too slow, we perform a greedy variant of this exhaustive search.

{\bf Greedy Search.}   
Initialize $\theta$, the set of selected (text,image) pairs, to the empty set. Then consider all candidate (text,image) pairs $q \in Q$ to add. For each, add it to the current selected set to create $\theta_q$ and evaluate its goodness by using $\theta_q$ to perform image retrieval over the set of annotated patches $D$ (where the score for image $i$ is given by the maximum score across all patches $p \in P_i$ from that image). Select the candidate pair $q^*$ that best increases image-retrieval accuracy (as measured by AUC of a precision-recall curve), grow $\theta := \theta_{q^*}$, and repeat. 


\textbf{Evaluation.} 
We evaluate PDC generated instruction pairs $\theta$ by using them as queries for an image retrieval task on a {\em held-out} test dataset. Intuitively, we reason that humans who see good categorical instructions should be able to retrieve images that contain the relevant categories, which is similar to a computational image retrieval task. Since we test on unbalanced, large datasets as seen in modern image retrieval settings, we only return the top 1000 unique retrievals. Thus, we also use modern retrieval metrics: precision, recall, and average precision(AP) at $k$. Note that precision at lower $k$ values is more important, as the most immediate visual images retrieved are the first accessed. Lastly, following Pascal VOC~\cite{everingham2010pascal}, we report per class AP and PR curves to accurately portray the results across classes with large discrepancies in samples. 

\section{Experiments}
In our experiments, we aim to answer two crucial questions: 1) Are the instruction pairs generated by PDC visually accurate, meaningful, and interesting? 2) Can we quantitatively show that our PDC instruction pairs are better than instruction pairs generated by methodological baselines? Additionally, we explore the query policies and functions that are and could have been part of PDC. Finally, we diagnose our generated instruction pairs in order to understand which aspects attribute to their success. 

\subsection{NuImages Performance Evaluation}
\label{sec:NPE}
\vspace{-0.2cm}
{\textbf{Implementation.}}
Our PDC experiments used NuImages, a 2D version of NuScenes~\cite{nuscenes}, as our dataset, in that it is one of the few datasets with released labeling instructions. Importantly this allows us to visually compare our generated instruction pairs with NuImages original pairs. NuImages contains 83,724 total images and 25 classes. However, we only evaluated 23 foreground classes as two classes are background classes ({\tt ego flat} and {\tt drivable surface}). As detailed in \cref{sec31}, we build our NuImages database index with approximately 14 million visual embeddings. When we query our database index, we set our FAISS index probe hyperparameter, indicating the number of clusters visited, to a high 300 to allow for high accuracy search. 

We select CLIP~\cite{radford2021learning} as our VLM as it is currently one of the best and largest VLM publicly available. Specifically, we use the ViT-B/32 pre-trained CLIP model. Following CLIP, when using category labels as text queries, we convert them into the following: ‘a photo of $\langle label\rangle$⟩’. PDC runs mainly on {\em only} CPU, except when we need to extract language/vision embeddings on GPU. Lastly, we run our PDC until a maximum of four pairs of instruction pairs are found. This is solely to speed up run time, as we find that AP improvement is marginal beyond four pairs. However, we emphasize that if qualitative results are favored, running PDC longer can surface more interesting results.

{\textbf{Evaluation.}}
To show statistical importance, we split our dataset into 5 folds, run PDC on 5 different splits, and test on 5 non-overlapping sets. Since some categories contain many samples, it is impractical to query $> 50\%$ of the dataset to achieve a recall of 1. Thus, when testing with (text, image) instruction pair(s), we retrieve 1,000 unique images for each category. Our PR curves, APs, mAPs are all calculated from exactly 1,000 returns. We average precision and recall at each $k$ up to 1,000 across folds.

\subsection{NuImages Performance Results}
\vspace{-0.2cm}
{\bf Baselines Comparisons.}
We first establish several baselines for comparison. Importantly, as with PDC, all of our baselines do not require model training. Our first two baselines are formed by taking aspects from the original NuImages instructions: 1) \textit{Original Texts} uses all given class labels/sub-class names/synonyms as single-modal text queries, which is CLIP's standard query. We extract these texts from the original text descriptions (e.g., `animal', ‘dog’, ‘cat’, ‘rat’ for class {\tt animal}) and combine the results of these text queries through \textit{Max-fusion}, detailed in~\cref{method:combine}. 2) \textit{Original Pairs} takes all image examples from the original instructions and pairs each image with a text from \textit{Original Texts} set (e.g., (`dog', $\langle$photo of dog$\rangle$)). Importantly, the images from \textit{Original Pairs} are manually selected by dataset curators. Thus, we consider \textit{Original Pairs} as a strong matching baseline.

We also include two baselines that rely on randomly selected examples: 1) \textit{Random BBoxes} randomly selects the same number of bounding box examples per class as present in our PDC final set. \textit{Random BBoxes} always selects the largest bounding box of the class in an image with multiple instances. 2) \textit{Random Pairs} uses the bboxes from \textit{Random Bboxes} and randomly pairs each bbox with text from Original Texts (e.g., (`dog', $\langle$photo of deer$\rangle$)). 

Finally, we include a mean shift (MS) baseline. Our MS baseline utilizes the training images to fit the model, in particularly using bboxes. Similar to \textit{Random BBoxes}, our MS baseline selects the largest bbox per class in an image. The closest training images to each final MS cluster centers are selected as the final visual examples. The examples with the class text are used as the final queries during evaluation on held-out set. MS parameters are default sklearn values.

\begin{table}[t!]

\caption[AP comparisons of instruction pairs generated by different methods]{Comparisons of instruction pairs generated by different methods. Average APs@1000 across 5 folds is displayed per class. Classes are sorted based on the number of images containing them. Note, most low performing classes have fewer examples or are ambiguous. Our method, PDC, performs the best for 21 of 23 classes.}

\label{tab:baseline} 
\addtolength{\tabcolsep}{-0.35em}
\resizebox{\linewidth}{!}{ 
\begin{tabular}{llccccc|c}
\toprule
Category      &\# Exs &Org. Ts &Org. Ps &Rnd. Bbs. &Rnd. Ps &MS &\textbf{PDC Pairs} \\
\midrule 
car&56517&$8.5^{\pm0.0}$&$5.7^{\pm0.0}$&$6.2^{\pm0.4}$&$4.6^{\pm0.4}$&$8.1^{\pm0.2}$&\bm{$8.8^{\pm0.1}$}\\
adult ped.&40241&$7.3^{\pm0.2}$&$2.6^{\pm0.0}$&$2.9^{\pm0.1}$&$3.0^{\pm0.2}$&$4.8^{\pm0.4}$&\bm{$12.0^{\pm0.2}$}\\
truck&23499&$12.5^{\pm0.3}$&$7.3^{\pm0.1}$&$2.7^{\pm0.2}$&$1.8^{\pm0.1}$&$3.1^{\pm0.3}$&\bm{$16.5^{\pm0.7}$}\\
traffic cone&22194&$20.7^{\pm0.1}$&$6.0^{\pm0.1}$&$15.0^{\pm1.6}$&$6.2^{\pm1.2}$&$16.3^{\pm1.3}$&\bm{$22.2^{\pm0.3}$}\\
traffic barrier&13607&$3.4^{\pm0.1}$&$0.5^{\pm0.0}$&$1.8^{\pm0.3}$&$0.8^{\pm0.1}$&$0.8^{\pm0.1}$&\bm{$16.9^{\pm2.3}$}\\
motorcycle&12523&$31.3^{\pm0.7}$&$0.6^{\pm0.0}$&$7.2^{\pm1.4}$&$2.3^{\pm0.6}$&$0.9^{\pm0.1}$&\bm{$33.6^{\pm1.4}$}\\
bicycle&11883&$24.6^{\pm1.1}$&$17.2^{\pm0.2}$&$7.6^{\pm1.7}$&$1.3^{\pm0.1}$&$5.9^{\pm0.7}$&\bm{$29.1^{\pm2.8}$}\\
rigid bus&7042&$14.4^{\pm1.1}$&$1.8^{\pm0.0}$&$1.9^{\pm0.3}$&$0.8^{\pm0.1}$&$0.7^{\pm0.1}$&\bm{$24.8^{\pm1.1}$}\\
construct. wrkr&5586&$26.2^{\pm2.1}$&$0.4^{\pm0.0}$&$3.9^{\pm0.6}$&$0.5^{\pm0.1}$&$6.8^{\pm1.0}$&\bm{$31.4^{\pm4.4}$}\\
construct. veh.&5258&$24.4^{\pm0.9}$&$1.4^{\pm0.1}$&$3.5^{\pm0.8}$&$1.8^{\pm0.5}$&$2.6^{\pm0.9}$&\bm{$39.1^{\pm1.0}$}\\
bicycle rack&2771&$9.4^{\pm1.4}$&$1.3^{\pm0.1}$&$3.7^{\pm0.7}$&$0.4^{\pm0.1}$&$1.0^{\pm1.0}$&\bm{$21.3^{\pm10.8}$}\\
push-pull obj.&2585&$1.2^{\pm0.2}$&$0.3^{\pm0.0}$&$0.4^{\pm0.0}$&$0.3^{\pm0.0}$&$0.2^{\pm0.1}$&\bm{$6.2^{\pm2.1}$}\\
trailer&2286&$1.9^{\pm0.2}$&$1.1^{\pm0.1}$&$3.8^{\pm1.1}$&$2.9^{\pm1.0}$&$1.3^{\pm0.3}$&\bm{$17.3^{\pm4.6}$}\\
debris&1840&$0.2^{\pm0.0}$&$0.3^{\pm0.0}$&$0.8^{\pm0.3}$&$0.5^{\pm0.2}$&$0.1^{\pm0.0}$&\bm{$8.9^{\pm1.6}$}\\
child ped.&1060&$1.2^{\pm0.2}$&$0.1^{\pm0.0}$&$0.5^{\pm0.1}$&$0.4^{\pm0.1}$&$0.2^{\pm0.2}$&\bm{$1.5^{\pm0.4}$}\\
pers. mobi. veh.&790&$2.7^{\pm0.3}$&$0.8^{\pm0.1}$&$0.4^{\pm0.1}$&$0.1^{\pm0.0}$&$0.2^{\pm0.3}$&\bm{$4.5^{\pm1.3}$}\\
police officer&356&$4.5^{\pm0.4}$&$1.1^{\pm0.3}$&$0.7^{\pm0.2}$&$0.1^{\pm0.0}$&$0.1^{\pm0.1}$&\bm{$5.5^{\pm5.2}$}\\
stroller&334&$3.8^{\pm0.9}$&\bm{$17.5^{\pm0.8}$}&$0.8^{\pm0.0}$&$0.1^{\pm0.0}$&$0.4^{\pm0.8}$&$16.5^{\pm3.9}$\\
animal&202&$0.0^{\pm0.0}$&$0.1^{\pm0.0}$&$0.0^{\pm0.0}$&$0.0^{\pm0.0}$&$0.0^{\pm0.0}$&\bm{$0.8^{\pm1.3}$}\\
bendy bus&169&$0.4^{\pm0.0}$&$0.0^{\pm0.0}$&$0.1^{\pm0.0}$&$0.0^{\pm0.0}$&$0.1^{\pm0.1}$&\bm{$6.9^{\pm4.4}$}\\
police vehicle&132&$0.8^{\pm0.1}$&$3.3^{\pm0.8}$&$0.8^{\pm0.2}$&$0.2^{\pm0.1}$&$0.2^{\pm0.2}$&\bm{$16.3^{\pm11.0}$}\\
ambulance&40&\bm{$0.5^{\pm0.1}$}&$0.1^{\pm0.0}$&$0.0^{\pm0.0}$&$0.0^{\pm0.0}$&$0.1^{\pm0.1}$&$0.1^{\pm0.2}$\\
wheelchair&33&$2.0^{\pm0.3}$&$1.2^{\pm0.2}$&$1.1^{\pm0.5}$&$0.0^{\pm0.0}$&$0.2^{\pm0.3}$&\bm{$14.7^{\pm9.2}$}\\
\hline
mAP &- & 8.77			&3.08 &2.86	&1.22	&2.35 &\textbf{15.44} \\
\bottomrule

\end{tabular}
}
\vspace{-0.3cm}
\end{table}

\textbf{NuImages Results.}
We first examine how PDC quantitatively compares with our baselines, as seen in ~\cref{tab:baseline}. PDC shows a significant improvement of 7.06 mAP over our strongest baseline, \textit{Original Texts}. Furthermore, PDC outperforms all baselines in 21 of 23 classes. In classes {\tt construction vehicle}, {\tt rigid bus}, and {\tt debris}, PDC outperforms the strongest contender by a large margin of 14.7, 10.4, 8.1AP respectively. We note that PDC improves on classes that are ambiguously defined, such as {\tt pushable pullable object} and {\tt debris}. In general, we see that classes with more instances in the dataset achieve higher AP, the exception being {\tt car}. PDC performs extremely well on {\tt car} with high precision and recall in the top 1000 retrievals. However, because there are so many ground truth {\tt car} instances, R@1000 is naturally low, leading to low AP.
\vspace{-0.1cm}

\begin{figure}[h!]
\centering
\caption{Average PR@1000 across 5 folds are shown for a subset of classes. Because each method retrieves 1000 samples, and classes are unbalanced, methods cannot reach the same or 1.0 recall. Full results in ~\cref{sup_nuimages} of the Appendix.}
\vspace{-.2cm}
\includegraphics[width=\linewidth]{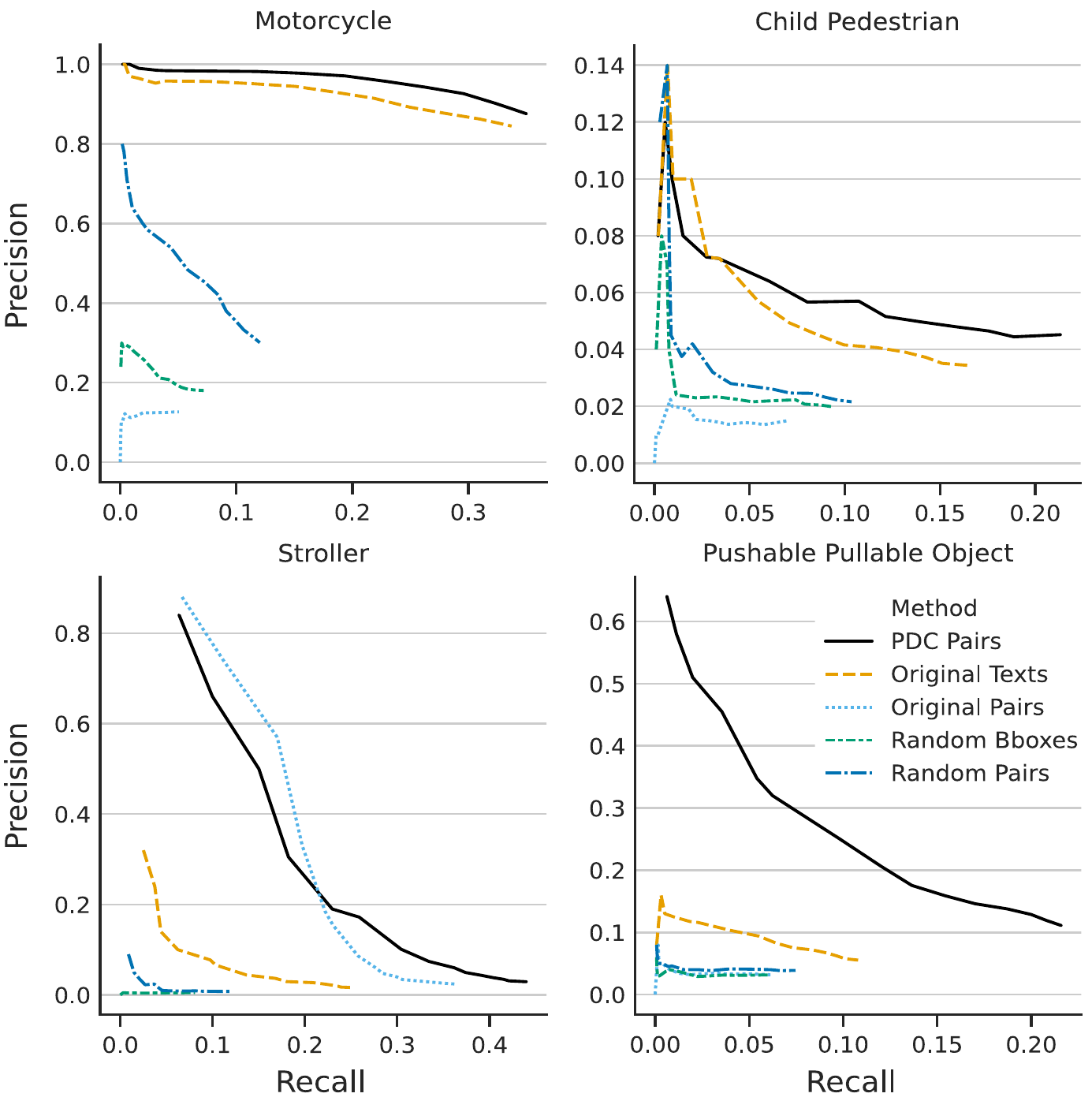}
\label{fig:pr} 
\vspace{-0.9cm}
\end{figure}
\begin{table}[ht!]
\vspace{-0.2cm}
\caption{COCO subset results: Comparisons of instruction pairs generated by different methods. Average APs across 5 folds is displayed per class. mAP is calculated across all 80 classes. PDC performs the best for 75 of 80 classes.}
\vspace{-.2cm}
\label{tab:main_coco} 
\addtolength{\tabcolsep}{-0.3em}
\resizebox{\linewidth}{!}{ 
\begin{tabular}{lcccc|c}
\toprule
Category      &\# Exs &Org. Ts &Rnd. Bbs. &Rnd. Ps &\textbf{PDC Pairs} \\
\midrule 
bottle&8880&$20.2^{\pm1.5}$&$0.3^{\pm0.3}$&$0.5^{\pm0.4}$& \bm{$31.0^{\pm2.1}$}\\
sports ball&4431&$43.9^{\pm2.4}$&$2.2^{\pm4.7}$&$4.5^{\pm9.5}$& \bm{$46.0^{\pm3.6}$}\\
skis&3202&$64.4^{\pm3.5}$&$4.4^{\pm5.8}$&$18.0^{\pm16.2}$& \bm{$80.2^{\pm1.8}$}\\
sheep&1594&$78.3^{\pm0.7}$&$40.6^{\pm25.2}$&$59.3^{\pm20.4}$& \bm{$85.4^{\pm1.4}$}\\
\hline
mAP &- & 42.0			&13.4 	&21.1	& \textbf{54.9} \\
\bottomrule
\end{tabular}
}
\vspace{-0.2cm}
\end{table}

\textbf{NuImages Further Investigation.}
We closely examine some of the most interesting classes' PR curves in ~\cref{fig:pr}. Looking at {\tt stroller}, one of two classes where \textit{Original Texts} outperforms PDC, we see that PDC still shows competitive results. While PDC achieves somewhat lower precision, it shows higher recall. For {\tt motorcycle}, we observe that while \textit{Original Texts} has both high precision and recall, PDC still outperforms it. In ambiguous classes such as {\tt pushable pullable object}, PDC manages to dramatically improve the precision. However, we note that classes such as {\tt child pedestrian} are challenging because adults are often mistaken for children.

\subsection{COCO Performance Evaluation}
\vspace{-0.2cm}
We also use PDC to generate LIs for COCO, a more class diverse detection benchmark. We compare COCO to the same baselines as NuImages: 1) \textit{Original Texts}; 2) \textit{Random BBoxes}; 3) \textit{Random Pairs}. However, as noted, we attempted but failed to obtain the original instructions from COCO curators. Thus, Original Pairs is not a viable baseline. For Original Texts, COCO contains fine-grain categories and, consequently, does not provide synonyms/subtypes. Thus, class names were used. PDC (54.9~mAP) outperforms its best competitor by a significant 12.9 points as shown in ~\cref{tab:main_coco}. APs, PR curves, and qualitative results for all COCO classes are reported in ~\cref{coco_app} of the Appendix. 

\subsection{Corner Cases and Prototypes}
\begin{figure}[h!]
\centering
\caption{PDC's generated instruction pairs. We show objects from diverse viewpoint, size, and type. NuImages: top 3 rows. COCO: bottom 2 rows. Complete results in ~\cref{sup_nuimages,coco_app} of the Appendix.}
\vspace{-0.2cm}
\includegraphics[width=\linewidth]{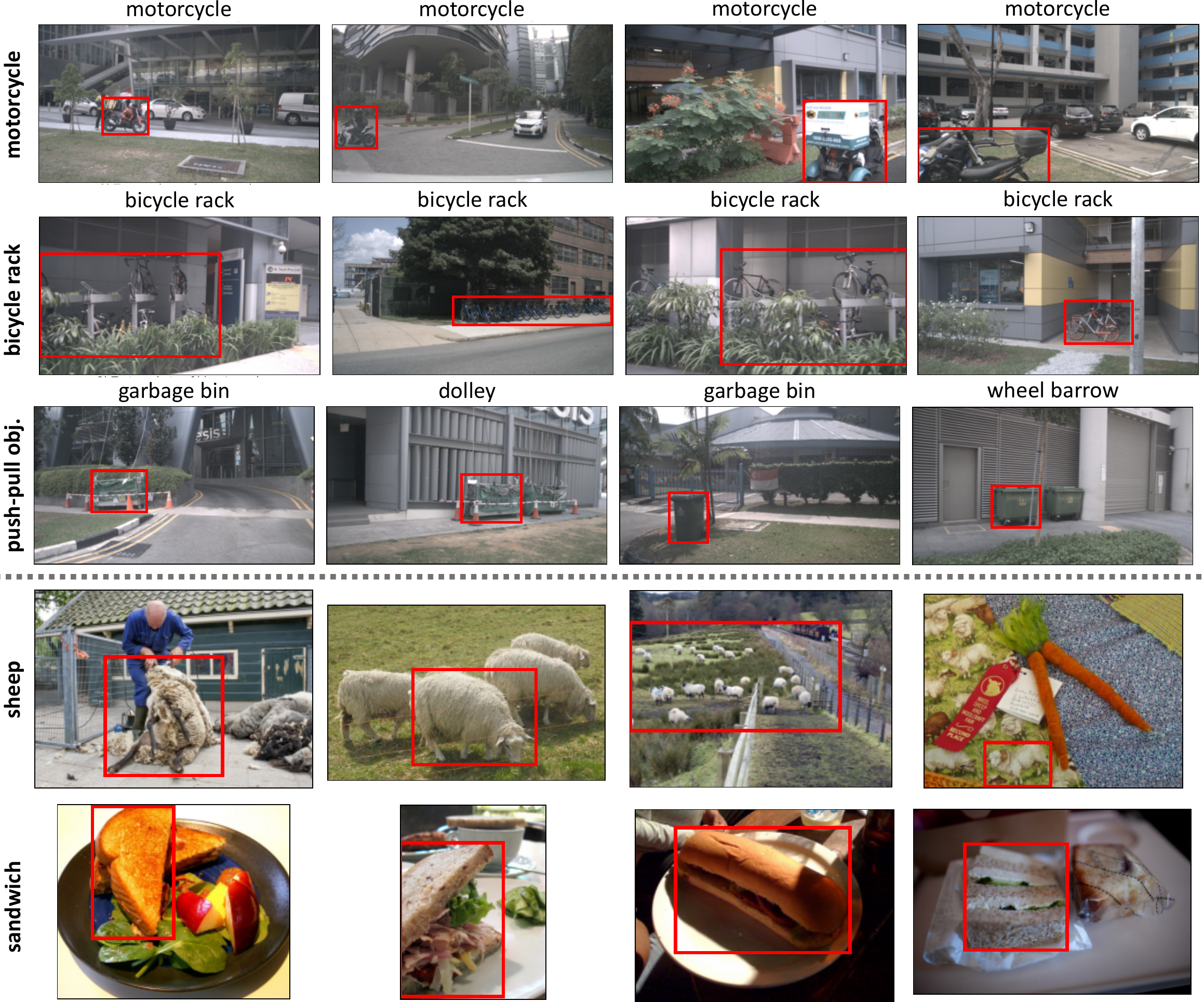}
\label{fig:qual} 
\vspace{-0.2cm}
\end{figure}

\vspace{-0.2cm}
We show the PDC generated instruction pairs in ~\cref{fig:qual}. All objects show diversity in viewpoints, sizes, and importantly sub-type. For NuImages, {\tt motorcycle} includes a corner case three-wheeled motorcycle and {\tt bicycle rack} includes a two tier rack. Note that some of these objects are partially occluded. For {\tt pushable pullable object}, PDC's instruction set conveys that most objects in the dataset are garbage bins variants. Although there is text to image mismatch, the errors are explainable (`wheel barrow' for a garbage bin with prominently shown wheels; `dolly' for a dolly-like garbage bin). For COCO,  {\tt sheep} includes a sheared sheep (corner), a sheep drawing (corner), and a real sheep (prototype). {\tt Sandwich} includes various triangle sandwiches and subs (corner). 

\newpage
\subsection{NuImages Human Behavioral Experiment}
\begin{figure}[h!]
\centering
\caption{A trial example for {\tt construction vehicle}. Correctly labeled objects are bounded by green bounding boxes. Participants are asked to pick set A or B.}
\vspace{-0.2cm}
\includegraphics[width=\linewidth]{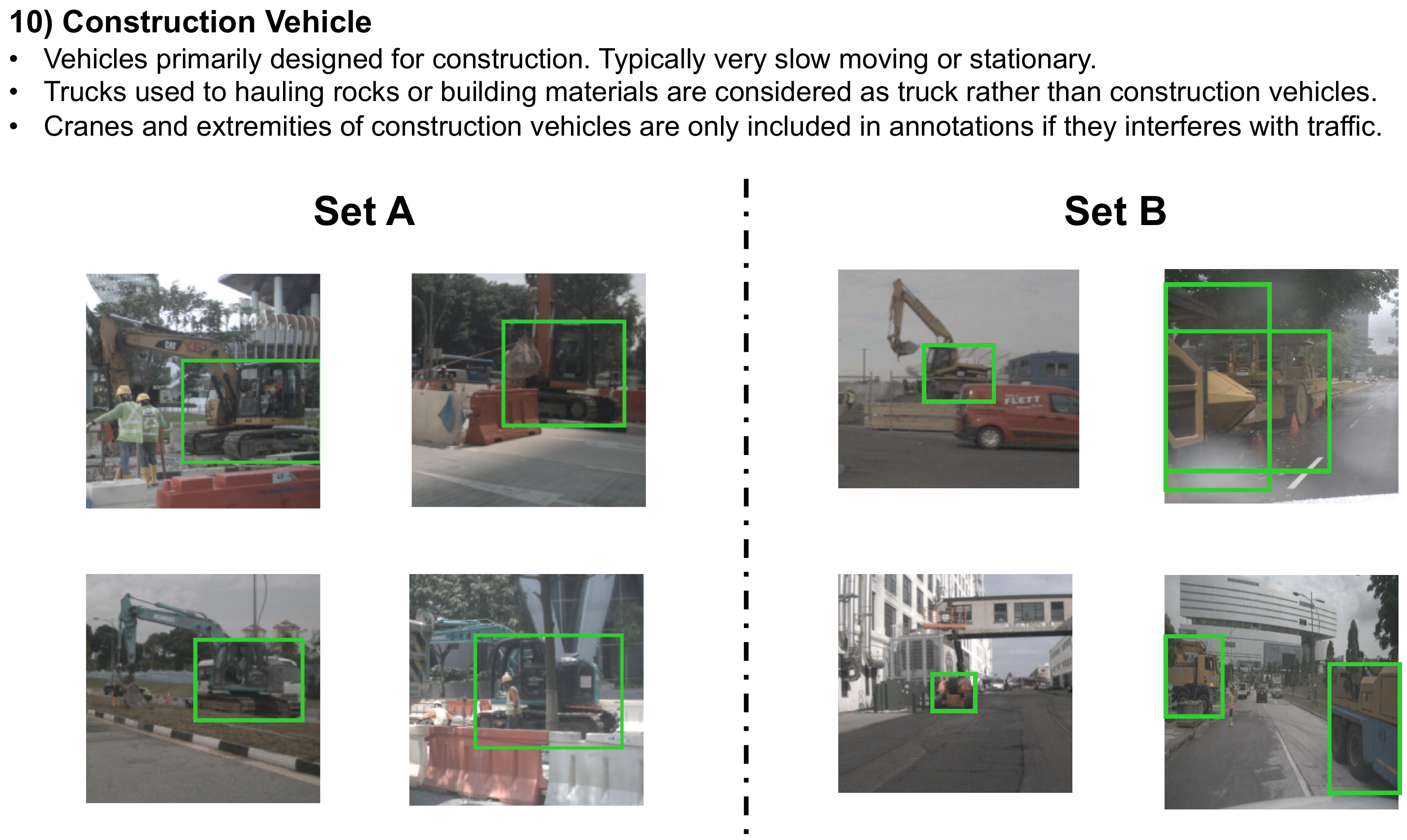}
\label{fig:study_slide}
\vspace{-0.4cm}
\end{figure}

\vspace{-0.2cm}
A human behavioral experiment evaluated how our generated NuImages instructions visually compare to the original NuImages instructions. The experiment consisted of 23 trials (one per category as in \ref{sec:NPE}). We show a sample trial in ~\cref{fig:study_slide} for class {\tt construction vehicle}. Each trial consists of the category name and category description provided by NuImages original instructions. Within each trial, there are two candidate image instruction sets: 1) images generated from PDC; 2) images from NuImages original instructions. All images contain a correct object encased with a bounding box. Participants select one of the two candidate instruction sets that they believe best guides them for future categorical annotation tasks (i.e., set A or B).

\begin{figure}[h!]
\centering
\caption{Participant responses for 23 NuImages categories are shown (N = 9). We observe that preferences for either original or PDC instructions are consistent across participants. Preferences are usually decided by a super majority of participants.}
\vspace{-0.2cm}
\includegraphics[width=\linewidth]{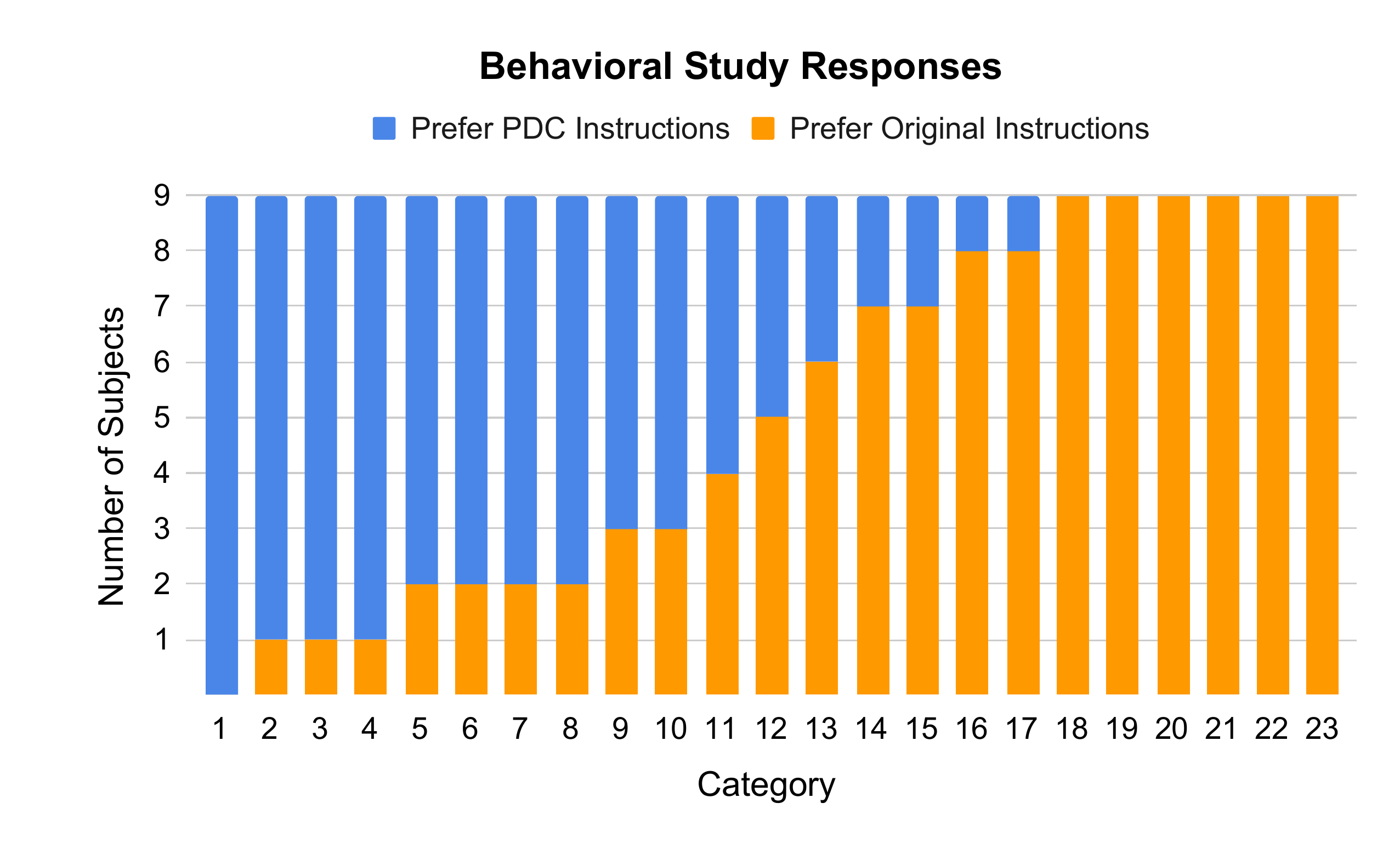}
\label{fig:study_results}
\vspace{-0.4cm}
\end{figure}

Complete behavioral study results are shown in ~\cref{fig:study_results}. Across all trials, the 9 participants preferred PDC generated instructions over the original instructions 44\% of the time. Because no direct inferences can be made from a statistically null effect (equal preference), we also examined whether participants were in agreement with one another for a given pair of instruction sets or were randomly choosing between generated and original instructions. Importantly participant preferences were not random, but were found to be consistent across participants with 78.8\% agreement for preferred PDC generated LIs and 87.8\% agreement for preferred original instructions. These results establish that, on the whole, generated and human selected instructions can both serve as visually effective annotation guides for potential future annotation tasks.

\begin{table}[h]
\caption{Comparisons of instruction pairs results fused by various query policies. Average APs@1000 across 5 folds is shown per class. Our final PDC setup gets the best mAP and outperforms the second best setup for 16 of 23 classes. Complete results in ~\cref{sup_nuimages} of the Appendix.}
\label{tab:main_diagA} 
\addtolength{\tabcolsep}{-0.2em}
\resizebox{\linewidth}{!}{ 
\begin{tabular}{lccccc|c}
\toprule
Category      &\# Exs &Sum:Avg &Early:Wt &Late:Naive & Late:Rank &\textbf{PDC Pairs} \\
\midrule 
truck&23499&$15.4^{\pm1.3}$&$16.5^{\pm1.2}$&$15.7^{\pm0.7}$&$16.1^{\pm0.6}$&\bm{$16.5^{\pm0.7}$}\\
construct. wrkr&5586&$27.6^{\pm4.3}$&\bm{$34.2^{\pm2.9}$}&$29.4^{\pm3.2}$&$31.4^{\pm3.5}$&$31.4^{\pm4.4}$\\
construct. veh.&5258&$36.5^{\pm2.4}$&$32.6^{\pm6.1}$&$30.7^{\pm2.5}$&$33.1^{\pm2.3}$&\bm{$39.1^{\pm1.0}$}\\
bicycle rack&2771&$17.7^{\pm10.6}$&$22.0^{\pm8.3}$&$16.7^{\pm6.7}$&$17.6^{\pm6.9}$&\bm{$21.3^{\pm10.8}$}\\
ambulance&40&$0.1^{\pm0.2}$&$0.1^{\pm0.1}$&$0.4^{\pm0.4}$&\bm{$0.4^{\pm0.4}$}&$0.1^{\pm0.2}$\\
\hline
mAP &- & 13.89			&14.75			&12.97			&13.69			&\textbf{15.44} \\
\bottomrule
\end{tabular}
}
\vspace{-0.2cm}
\end{table}

\textbf{Query Fusion Policy Ablations}
 We explore various ways to query fusion policies as described in ~\cref{sec32,method:combine}. In PDC, we use \textit{Early-Fusion: Sum} and Max-Fusion as the query policy and method to combine results across all generated instruction sets respectively. Given the same PDC generated instruction sets, our ablation results on different methods to query with and combine results are shown in ~\cref{tab:main_diagA}. Results illustrate that our final setup outperforms the next best, \textit{Early-Fusion: Weighted}, by 0.69 mAP. We observe that all Early-Fusion methods--\textit{Sum:Avg}, \textit{Weighted}, \textit{Sum:Max}(used by PDC)--outperforms all Late-Fusion methods (\textit{Naive}, \textit{Inverse Rank}). By combining queries to create a single query, Early-Fusion methods are computationally faster and cheaper than Late-Fusion methods.


\begin{table}[h]
\caption{Results from using only the texts or bboxes of our PDC instruction set. Average APs@1000 across 5 folds is displayed per class. Using both text and bboxes provides the best APs for 17 of 23 classes. Complete results in ~\cref{sup_nuimages} of the Appendix.}
\label{tab:main_diagB} 
\addtolength{\tabcolsep}{-0.2em}
\resizebox{\linewidth}{!}{ 
\begin{tabular}{lccc|c}
\toprule
Category      &\# Exs &Texts &Bboxes   &\textbf{PDC Pairs} \\
\midrule 
bicycle&11883&\bm{$30.4^{\pm0.1}$}&$23.5^{\pm1.3}$&$29.1^{\pm2.8}$\\
construction vehicle&5258&$20.1^{\pm0.6}$&$38.6^{\pm0.5}$&\bm{$39.1^{\pm1.0}$}\\
bendy bus&169&$0.9^{\pm0.1}$&$8.6^{\pm0.9}$&\bm{$6.9^{\pm4.4}$}\\
police vehicle&132&$4.7^{\pm1.1}$&$18.9^{\pm1.9}$&$16.3^{\pm11.0}$\\
\hline
mAP &- 	&10.50			&13.56			&\textbf{15.44} \\
\bottomrule
\end{tabular}
}
\end{table}
\textbf{Generated Instruction Pairs Diagnostics}
Finally, we examine which aspect (text or bbox) of PDC generated instruction pairs contributes the most to our final results. As observed in~\cref{tab:main_diagB}, PDC text and bbox pairs outperforms the next best, bbox only, by 1.88 mAP. In general, using only bboxes is better than using only texts. We show our largest improvement (+18.95~AP in {\tt construction vehicle}) over only texts and worst decrements (-2.67~AP in {\tt police vehicle}) over only bboxes.

\section{Conclusion}
Detailed and clear annotation policies are integral to large scale dataset creation which, in turn, forms the backbone for much of modern deep learning. Yet few datasets include annotation instructions. This omission presents a challenge for dataset transparency, reproducibility, and error interpretation. To address this gap, we propose a new task, Labeling Instruction Generation (LIG), and a fast and computationally efficient \textit{post-hoc} solution - Proxy Dataset Curator (PDC) - to LIG that serves as a substitute or enhancement for dataset curators. PDC can efficiently, and without model training, replicate the laborious manual iterative process of instruction policy refinement and outperforms our strongest baselines by a significant margin. Future work should continue to explore solutions to LIG that may provide better refined and well-specified annotation instructions.

\section{Acknowledgements}
I would like to sincerely thank Jayanth Koushik for various discussions, comments, advice, and suggestions for this project. This work was supported by National Science Foundation Graduate Research Program Fellowship (DGE1745016) and CMU Argo AI Center for Autonomous Vehicle Research. 

{\small
\bibliographystyle{ieee_fullname}
\bibliography{egbib}

\begin{thebibliography}{10}\itemsep=-1pt

\bibitem{antol2015vqa}
Stanislaw Antol, Aishwarya Agrawal, Jiasen Lu, Margaret Mitchell, Dhruv Batra,
  C~Lawrence Zitnick, and Devi Parikh.
\newblock Vqa: Visual question answering.
\newblock In {\em Proceedings of the IEEE international conference on computer
  vision}, pages 2425--2433, 2015.

\bibitem{bearman2016s}
Amy Bearman, Olga Russakovsky, Vittorio Ferrari, and Li Fei-Fei.
\newblock What’s the point: Semantic segmentation with point supervision.
\newblock In {\em European conference on computer vision}, pages 549--565.
  Springer, 2016.

\bibitem{beckwith2021wordnet}
Richard Beckwith, Christiane Fellbaum, Derek Gross, and George~A Miller.
\newblock Wordnet: A lexical database organized on psycholinguistic principles.
\newblock In {\em Lexical acquisition: Exploiting on-line resources to build a
  lexicon}, pages 211--232. Psychology Press, 2021.

\bibitem{birhane2021large}
Abeba Birhane and Vinay~Uday Prabhu.
\newblock Large image datasets: A pyrrhic win for computer vision?
\newblock In {\em 2021 IEEE Winter Conference on Applications of Computer
  Vision (WACV)}, pages 1536--1546. IEEE, 2021.

\bibitem{nuscenes}
Holger Caesar, Varun Bankiti, Alex~H. Lang, Sourabh Vora, Venice~Erin Liong,
  Qiang Xu, Anush Krishnan, Yu Pan, Giancarlo Baldan, and Oscar Beijbom.
\newblock nuscenes: A multimodal dataset for autonomous driving.
\newblock In {\em CVPR}, 2020.

\bibitem{clinchant2011semantic}
S. Clinchant, Ah-Pine J., and G. Csurka.
\newblock Semantic combination of textual and visual information in multimedia
  retrieval.
\newblock In {\em ACM ICMR}, pages 1--8, 2011.

\bibitem{cohen2022my}
Niv Cohen, Rinon Gal, Eli~A Meirom, Gal Chechik, and Yuval Atzmon.
\newblock " this is my unicorn, fluffy": Personalizing frozen vision-language
  representations.
\newblock {\em arXiv preprint arXiv:2204.01694}, 2022.

\bibitem{Craswell2009}
Nick Craswell.
\newblock {\em Mean Reciprocal Rank}, pages 1703--1703.
\newblock Springer US, Boston, MA, 2009.

\bibitem{deng2009imagenet}
Jia Deng, Wei Dong, Richard Socher, Li-Jia Li, Kai Li, and Li Fei-Fei.
\newblock Imagenet: A large-scale hierarchical image database.
\newblock In {\em CVPR}, 2009.

\bibitem{everingham2010pascal}
Mark Everingham, Luc Van~Gool, Christopher~KI Williams, John Winn, and Andrew
  Zisserman.
\newblock The pascal visual object classes (voc) challenge.
\newblock {\em IJCV}, 88(2):303--338, 2010.

\bibitem{gebru2021datasheets}
T. Gebru et~al.
\newblock Datasheets for datasets.
\newblock {\em ACM}, 2021.

\bibitem{goyal2017making}
Yash Goyal, Tejas Khot, Douglas Summers-Stay, Dhruv Batra, and Devi Parikh.
\newblock Making the v in vqa matter: Elevating the role of image understanding
  in visual question answering.
\newblock In {\em Proceedings of the IEEE conference on computer vision and
  pattern recognition}, pages 6904--6913, 2017.

\bibitem{gupta2019lvis}
Agrim Gupta, Piotr Dollar, and Ross Girshick.
\newblock Lvis: A dataset for large vocabulary instance segmentation.
\newblock In {\em CVPR}, pages 5356--5364, 2019.

\bibitem{Itri2018}
J.~N. Itri et~al.
\newblock Heuristics and cognitive error in medical imaging.
\newblock {\em AJR Am J Roentgenol}, 210(5):1097--1105, 2018.

\bibitem{jain2013predicting}
Suyog~Dutt Jain and Kristen Grauman.
\newblock Predicting sufficient annotation strength for interactive foreground
  segmentation.
\newblock In {\em Proceedings of the IEEE International Conference on Computer
  Vision}, pages 1313--1320, 2013.

\bibitem{jia2021scaling}
Chao Jia, Yinfei Yang, Ye Xia, Yi-Ting Chen, Zarana Parekh, Hieu Pham, Quoc Le,
  Yun-Hsuan Sung, Zhen Li, and Tom Duerig.
\newblock Scaling up visual and vision-language representation learning with
  noisy text supervision.
\newblock In {\em International Conference on Machine Learning}, pages
  4904--4916. PMLR, 2021.

\bibitem{johnson2019billion}
Jeff Johnson, Matthijs Douze, and Herv{\'e} J{\'e}gou.
\newblock Billion-scale similarity search with {GPUs}.
\newblock {\em IEEE Transactions on Big Data}, 7(3):535--547, 2019.

\bibitem{johnson2015image}
Justin Johnson, Ranjay Krishna, Michael Stark, Li-Jia Li, David Shamma, Michael
  Bernstein, and Li Fei-Fei.
\newblock Image retrieval using scene graphs.
\newblock In {\em Proceedings of the IEEE conference on computer vision and
  pattern recognition}, pages 3668--3678, 2015.

\bibitem{kang2022}
Daniel Kang.
\newblock {ML} models are increasingly being deployed in mission-critical
  settings.
\newblock {\em Online Communication, LinkedIn}, Sept 2022.

\bibitem{kangfinding}
Daniel. Kang et~al.
\newblock Finding label errors in autonomous vehicle data with learned
  observation assertions.
\newblock 2022.

\bibitem{kovashka2015discovering}
Adriana Kovashka and Kristen Grauman.
\newblock Discovering attribute shades of meaning with the crowd.
\newblock {\em International Journal of Computer Vision}, 114(1):56--73, 2015.

\bibitem{kovashka2011actively}
Adriana Kovashka, Sudheendra Vijayanarasimhan, and Kristen Grauman.
\newblock Actively selecting annotations among objects and attributes.
\newblock In {\em ICCV}, pages 1403--1410. IEEE, 2011.

\bibitem{krasin2017openimages}
Ivan Krasin, Tom Duerig, Neil Alldrin, Vittorio Ferrari, Sami Abu-El-Haija,
  Alina Kuznetsova, Hassan Rom, Jasper Uijlings, Stefan Popov, Andreas Veit,
  et~al.
\newblock Openimages: A public dataset for large-scale multi-label and
  multi-class image classification.
\newblock {\em Dataset available from https://github. com/openimages}, 2(3):18,
  2017.

\bibitem{openimages}
Ivan Krasin, Tom Duerig, Neil Alldrin, Andreas Veit, Sami Abu-El-Haija, Serge
  Belongie, David Cai, Zheyun Feng, Vittorio Ferrari, Victor Gomes, Abhinav
  Gupta, Dhyanesh Narayanan, Chen Sun, Gal Chechik, and Kevin Murphy.
\newblock Openimages: A public dataset for large-scale multi-label and
  multi-class image classification.
\newblock {\em Dataset available from https://github.com/openimages}, 2016.

\bibitem{kumar2022fine}
Ananya Kumar, Aditi Raghunathan, Robbie Jones, Tengyu Ma, and Percy Liang.
\newblock Fine-tuning can distort pretrained features and underperform
  out-of-distribution.
\newblock {\em arXiv preprint arXiv:2202.10054}, 2022.

\bibitem{li2022grounded}
Liunian~Harold Li, Pengchuan Zhang, Haotian Zhang, Jianwei Yang, Chunyuan Li,
  Yiwu Zhong, Lijuan Wang, Lu Yuan, Lei Zhang, Jenq-Neng Hwang, et~al.
\newblock Grounded language-image pre-training.
\newblock In {\em Proceedings of the IEEE/CVF Conference on Computer Vision and
  Pattern Recognition}, pages 10965--10975, 2022.

\bibitem{lin2014microsoft}
Tsung-Yi Lin, Michael Maire, Serge Belongie, James Hays, Pietro Perona, Deva
  Ramanan, Piotr Doll{\'a}r, and C~Lawrence Zitnick.
\newblock Microsoft coco: Common objects in context.
\newblock In {\em ECCV}, pages 740--755. Springer, 2014.

\bibitem{liu2021image}
Zheyuan Liu, Cristian Rodriguez-Opazo, Damien Teney, and Stephen Gould.
\newblock Image retrieval on real-life images with pre-trained
  vision-and-language models.
\newblock In {\em Proceedings of the IEEE/CVF International Conference on
  Computer Vision}, pages 2125--2134, 2021.

\bibitem{lu202012}
Jiasen Lu, Vedanuj Goswami, Marcus Rohrbach, Devi Parikh, and Stefan Lee.
\newblock 12-in-1: Multi-task vision and language representation learning.
\newblock In {\em Proceedings of the IEEE/CVF Conference on Computer Vision and
  Pattern Recognition}, pages 10437--10446, 2020.

\bibitem{marino2019ok}
Kenneth Marino, Mohammad Rastegari, Ali Farhadi, and Roozbeh Mottaghi.
\newblock Ok-vqa: A visual question answering benchmark requiring external
  knowledge.
\newblock In {\em Proceedings of the IEEE/CVF Conference on Computer Vision and
  Pattern Recognition}, pages 3195--3204, 2019.

\bibitem{radford2021learning}
Alec Radford, Jong~Wook Kim, Chris Hallacy, Aditya Ramesh, Gabriel Goh,
  Sandhini Agarwal, Girish Sastry, Amanda Askell, Pamela Mishkin, Jack Clark,
  et~al.
\newblock Learning transferable visual models from natural language
  supervision.
\newblock In {\em ICML}, pages 8748--8763. PMLR, 2021.

\bibitem{recht2019imagenet}
B. Recht et~al.
\newblock Do {ImageNet} classifiers generalize to {ImageNet}?
\newblock In {\em ICML}, pages 5389--5400. PMLR, 2019.

\bibitem{russakovsky2015best}
Olga Russakovsky, Li-Jia Li, and Li Fei-Fei.
\newblock Best of both worlds: human-machine collaboration for object
  annotation.
\newblock In {\em CVPR}, pages 2121--2131, 2015.

\bibitem{skantze2022collie}
Gabriel Skantze and Bram Willemsen.
\newblock Collie: Continual learning of language grounding from language-image
  embeddings.
\newblock {\em Journal of Artificial Intelligence Research}, 74:1201--1223,
  2022.

\bibitem{vo2019composing}
Nam Vo, Lu Jiang, Chen Sun, Kevin Murphy, Li-Jia Li, Li Fei-Fei, and James
  Hays.
\newblock Composing text and image for image retrieval-an empirical odyssey.
\newblock In {\em Proceedings of the IEEE/CVF conference on computer vision and
  pattern recognition}, pages 6439--6448, 2019.

\bibitem{Wilkinson2016}
M.~D. Wilkinson et~al.
\newblock {The FAIR Guiding Principles for scientific data management and
  stewardship}.
\newblock {\em Sci Data}, 2016.

\bibitem{wortsman2022robust}
Mitchell Wortsman, Gabriel Ilharco, Jong~Wook Kim, Mike Li, Simon Kornblith,
  Rebecca Roelofs, Raphael~Gontijo Lopes, Hannaneh Hajishirzi, Ali Farhadi,
  Hongseok Namkoong, et~al.
\newblock Robust fine-tuning of zero-shot models.
\newblock In {\em Proceedings of the IEEE/CVF Conference on Computer Vision and
  Pattern Recognition}, pages 7959--7971, 2022.

\bibitem{xu2017scene}
Danfei Xu, Yuke Zhu, Christopher~B Choy, and Li Fei-Fei.
\newblock Scene graph generation by iterative message passing.
\newblock In {\em Proceedings of the IEEE conference on computer vision and
  pattern recognition}, pages 5410--5419, 2017.

\bibitem{yang2018graph}
Jianwei Yang, Jiasen Lu, Stefan Lee, Dhruv Batra, and Devi Parikh.
\newblock Graph r-cnn for scene graph generation.
\newblock In {\em Proceedings of the European conference on computer vision
  (ECCV)}, pages 670--685, 2018.

\bibitem{yuan2021florence}
Lu Yuan, Dongdong Chen, Yi-Ling Chen, Noel Codella, Xiyang Dai, Jianfeng Gao,
  Houdong Hu, Xuedong Huang, Boxin Li, Chunyuan Li, et~al.
\newblock Florence: A new foundation model for computer vision.
\newblock {\em arXiv preprint arXiv:2111.11432}, 2021.

\bibitem{zhou2017scene}
Bolei Zhou, Hang Zhao, Xavier Puig, Sanja Fidler, Adela Barriuso, and Antonio
  Torralba.
\newblock Scene parsing through ade20k dataset.
\newblock In {\em Proceedings of the IEEE conference on computer vision and
  pattern recognition}, pages 633--641, 2017.

\bibitem{zhou2021learning}
Kaiyang Zhou, Jingkang Yang, Chen~Change Loy, and Ziwei Liu.
\newblock Learning to prompt for vision-language models.
\newblock {\em arXiv preprint arXiv:2109.01134}, 2021.

\end{thebibliography}
}

\clearpage

\setcounter{section}{0}
\renewcommand*{\thesection}{\Alph{section}}
\setcounter{figure}{0}
\renewcommand{\figurename}{\bfseries Figure}
\renewcommand{\thefigure}{A\arabic{figure}}

\setcounter{table}{0}
\renewcommand{\tablename}{\bfseries Table}
\renewcommand{\thetable}{A\arabic{table}}

\section{Pseudocode}
\vspace{-0.2cm}
\label{code_app}
\textbf{Notation.}
Let $I$ be a set of images. For each $i \in I$, let $B_i$ be a set of bounding boxes, $b_i^c$. A bounding box $b_i^c$ is said to be labeled with category $c \in C$. We denote our image encoder taken from our V\&L model as $f_v^{en}: I \to \mathbb{R}^d$, mapping images to vectors in $\mathbb{R}^d$. 

For each image $i$, let $P_i \subset I$ be a set of image patches. We define the database $D:=\{ f_v^{en}(p): p \in P_i, i \in I\}$ as the embedding of all patches. For our experiments, we define two database indexes, $D_{tr}$ and $D_{te}$, built from training $I_{tr}$ and evaluated on $I_{te}$ image sets respectively.

Finally, we let $T$ be the set of words. For each category $c \in C$, let $L_c \subseteq T$ be the words that are associated with $c$. We denote our text encoder taken from our V\&L model as $f_t^{en}: T \to \mathbb{R}^d$, mapping words to vectors in $\mathbb{R}^d$.

\begin{algorithm}
\caption{PDC Framework} 
\label{algo}
\begin{algorithmic}[1]
\Require Category, $c$; the words associated with $c$, $L_c$; the training index $D_{tr}$
\Ensure $\Theta$, a set of instruction pair(s) $(b_i^c,t)$, where $b_i^c \in B_i$ and $t \in T$
\State Let $I^c = \{ i \in I_{tr}: \exists b_i^c \in B_i\}$ \label{op1}
\State Let $V_c \{\argmax_{b_i^c \in B_i}  \mathrm{area}(b_i^c): i \in I^c\}$ \label{op2}
\State Define $r_{tv}: V_c \times L_c \times D_{tr}$
\newline
$r_{tv}(b,w,i) = \mathrm{PatchFus}_{p \in P_i}($\newline
    \hspace*{5em} $\mathrm{MultiScore}(f_v^{en}(b), f_t^{en}(w), f_v^{en}(p)) )$ \label{op3} 
\State Define $J_{tv}(b,w) = \mathrm{topk}_{i \in I_{tr}} r_{tv}(b,w,i)$ \label{op4}
\State Define $\alpha_{tv}(b,w) = \mathrm{AUC}(\{r_{tv}(b,w,i): i \in J_{tv}\}$ \label{op5}
\State Let $(b^\star,w^\star) = \argmax_{(b,w) \in V_t \times L_c} \alpha_{tv}(b,w)$ \label{op6}
\State Add $(b^\star,w^\star)$ to $\Theta$ \label{op7}
\State Define $s^*(\Theta):= $\newline 
\hspace*{3em} $\mathrm{AUC}(\bigcup_{(b,w) \in \Theta} \{ r_{tv}(b,w,i): i \in J_{tv}(b,w)\})$ \label{op8}
\State Let $\Theta^\prime = \Theta^\star$
\While{$s^*(\Theta^\prime) \geq s^*(\Theta^\star)$}   
    \State $\Theta^\star = \Theta^\prime$
    \State $(b^\prime, w^\prime) = \argmax_{(b,w) \in V_t \times L_c} s^\star(\Theta^\star \bigcup \{b,w\})$
    \State $\Theta^\prime = \Theta^\star \bigcup \{b,w\}$
\EndWhile \label{op14}
\end{algorithmic}
\end{algorithm}
\vspace{-0.2cm}
{\bf Algorithm.}
We formally describe our PDC algorithm as depicted in ~\cref{algo}. ~\cref{algo} runs on all $c \in C$. We assume that for each $c$, there exists a set of associated descriptive words. For example, class {\tt animal} has the following labels/subclasses/synonyms as the set of words, (`animal', `dog', `bird', `cat'). First, in ~\cref{op1,op2} we define $V_c$ as the set of the largest bounding box object labeled $c$ for each image in our training set $I_{tr}$. Now, we have a potential pool of texts ($L_c$) and images ($V_c$). From this pool, we will be matching one text and one image to create pairs that will compose our final instruction pairs. 

In the following process, we match every text to every image to create every potential (text, image) pair, $(b, w)$. For each potential pair, in ~\cref{op3}, we utilize them as a multi-modal query against our index of patches. Here, we intuitively measure each pair's effectiveness as an instruction pair by measuring its capabilities at image retrieval. In ~\cref{op4,op5}, we utilize \textit{PatchFusion} to retrieve a set of top $k$ image scores indicating highest similarity to each multi-modal query. In image retrieval, it is common practice to use precision-recall at $k$ as a metric. Similarly, from a set of image scores, we can measure our precision at several $k$ steps. With a precision-recall at $k$ curve, we can measure its area under the curve (AUC). The higher the AUC, the better our query is. Thus, we measure the AUC for each set of results from each of our potential (text, image) pair, $(b,w)$. In ~\cref{op6,op7}, we add the pair with best AUC into our final output instruction set.

With the first instruction pair decided, PDC grows the instruction set as much as possible in ~\cref{op8} to ~\cref{op14}. To do so, we continue to comb through our potential text($L_c$) and bounding box($V_c$) pool and create all potential pairs. While the AUC of our final outcome instruction set is still improving, we continuously add a potential new pair into our outcome set and test its new AUC. If this new pair improves the outcome's AUC, we add it into the outcome set. 

\vspace{-0.2cm}
\section{NuImages Results and Ablations}
\label{sup_nuimages}
\vspace{-0.1cm}

\begin{figure*}[h!]
\centering
\caption{NuImages results: PDC's generated NuImages instruction pairs are shown. In these sets, we observe different subtypes of classes, objects in different sizes and viewpoints, and various text synonyms. In `shuttle', we see a shuttle (2nd image) that is correctly paired with the text `bendy shuttle'. In `truck', we see various types of trucks/lorries. However, we see a mismatched dump truck that is also paired with the text `pickup truck'. In `adult pedestrian' and `construction worker', we can observe people in various outfits, locations, and positions - a person sitting in the right most image of adult pedestrian. }
\includegraphics[width=\linewidth]{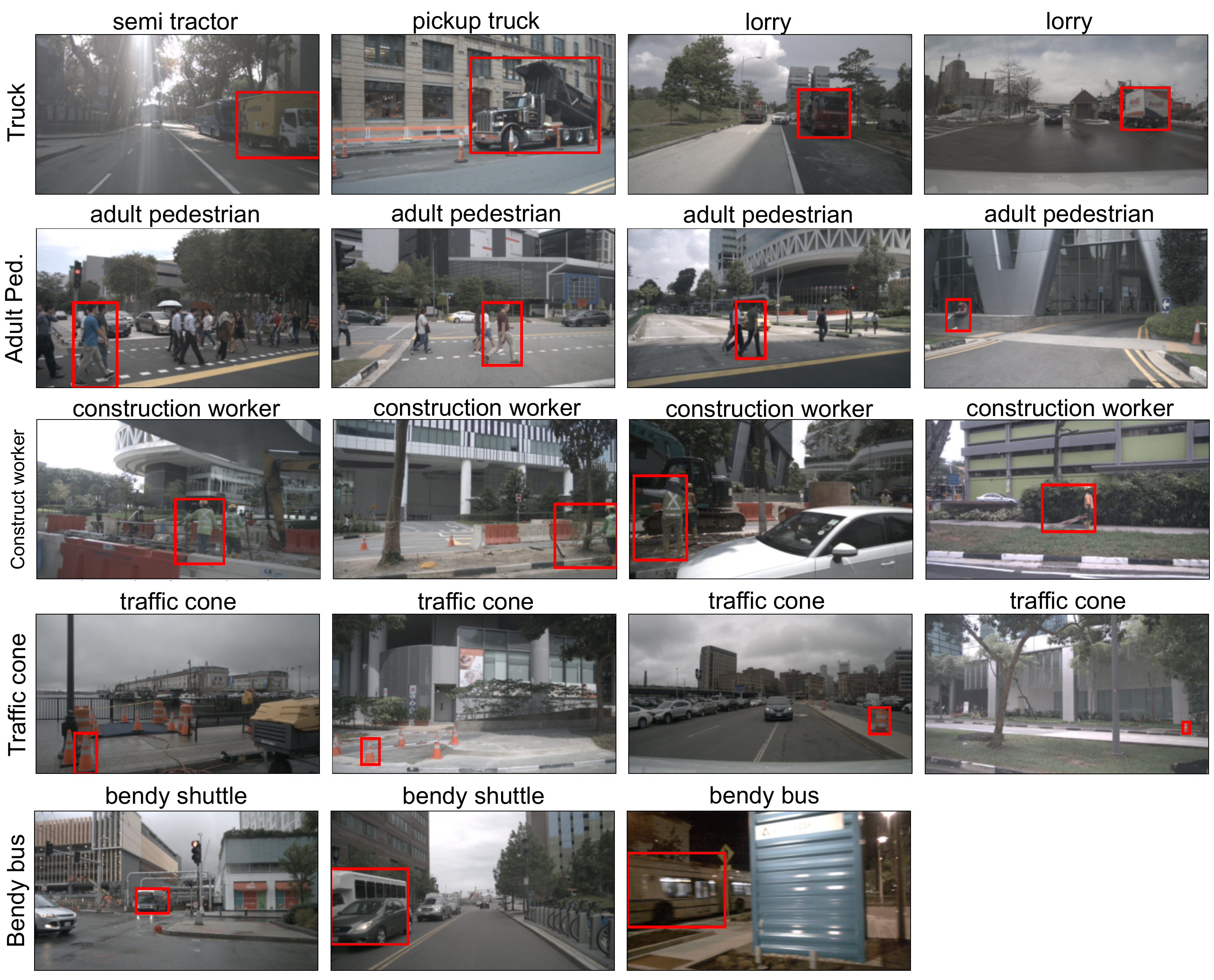}
\label{fig:nuimages_qual} 
\vspace{-0.2cm}
\end{figure*}

\textbf{PDC Generated Instruction Pairs}
Additionally, we show a subset of our generated instruction pairs for NuImages in ~\cref{fig:nuimages_qual} since we have too many qualitative results that can reasonably be displayed in a paper.

\begin{figure*}[ht!]
\centering
\caption{NuImages Baselines results: We display the per class AP curve across 5 folds here in parts. These plots correspond to the calculated AP in Tabs. 1 and subset PR curves in Fig. 4 in the main paper. The solid black line is our PDC Pair. We note that for each class, our PDC curve is comfortably above the others. In particular, we see significant increases in harder classes that achieves low precisions with other baselines (e.g. `debris', `pushable pullable object', `trailer', `police vehicle', `bendy bus'). Lastly, classes with strong performance using \textit{Original Texts} baselines still see a noticeable improvement with our PDC framework.}
\includegraphics[width=\linewidth]{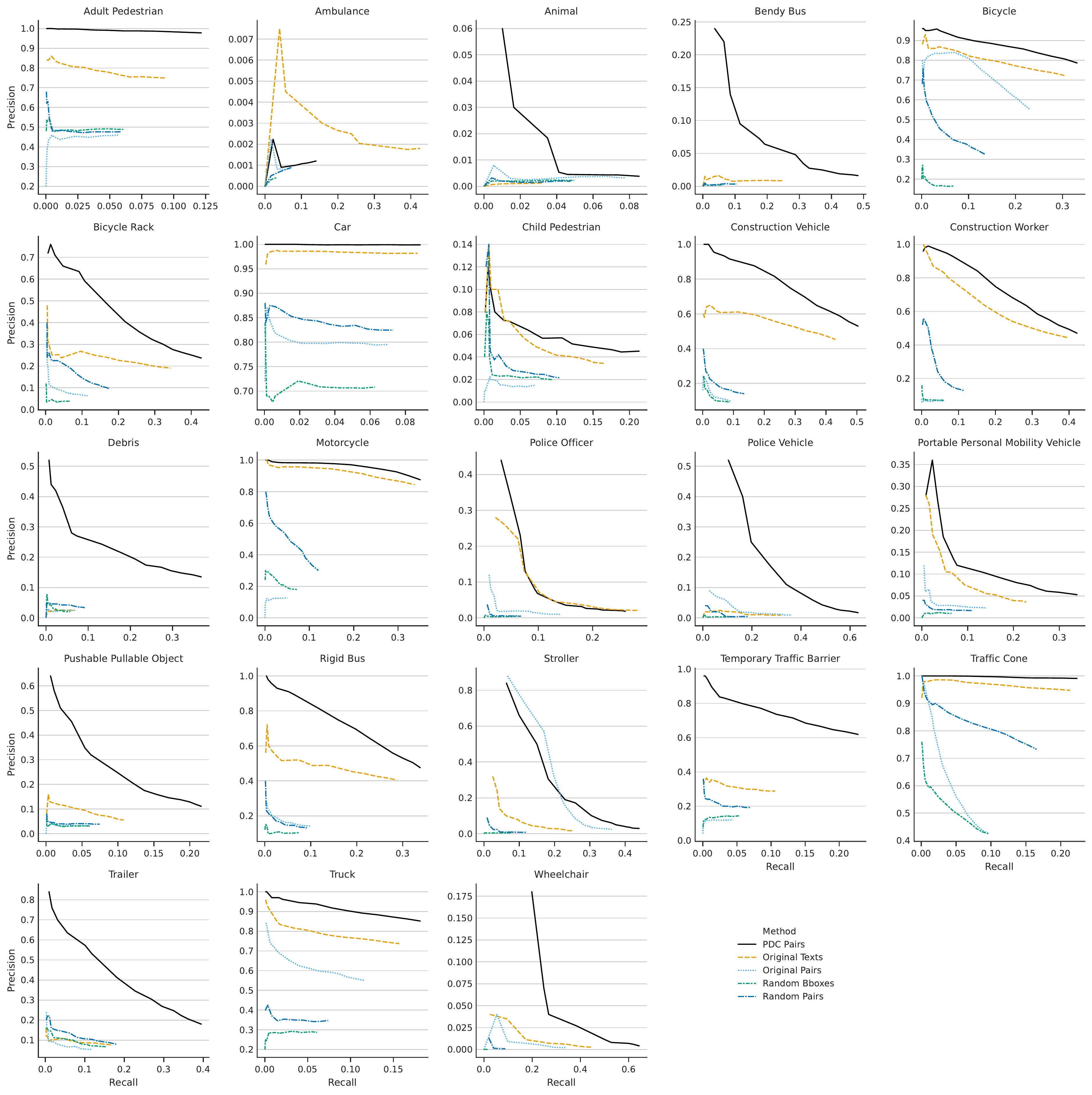}
\label{fig:baseline_sup}
\end{figure*}

\textbf{Class PR curves.} In addition to Fig. 5 (main), we provide class PR curves for all NuImages classes in ~\cref{fig:baseline_sup}. 

\begin{figure*}[ht!]
\centering
\caption{NuImages Query Fusion Policy Ablations results: We display the per class AP curve across 5 folds here in parts. These plots correspond to the calculated AP in Table 2 in the main paper. The solid black line is our PDC Pair. The PR curves indicate two main points: 1) \textit{Late Fusion Naive} and \textit{Inverse Rank} both underperform in almost all categories, 2) Our Early Fusion methods show similar results. Thus, we rely on AP to reliably inform us that \textit{Early Fusion: Sum, Max} is our best fusion method across multiple queries. In the main paper, we see that our final PDC outperforms the next best (\textit{Early Fusion: Weighted}) by 0.69 mAP.}
\includegraphics[width=\linewidth]{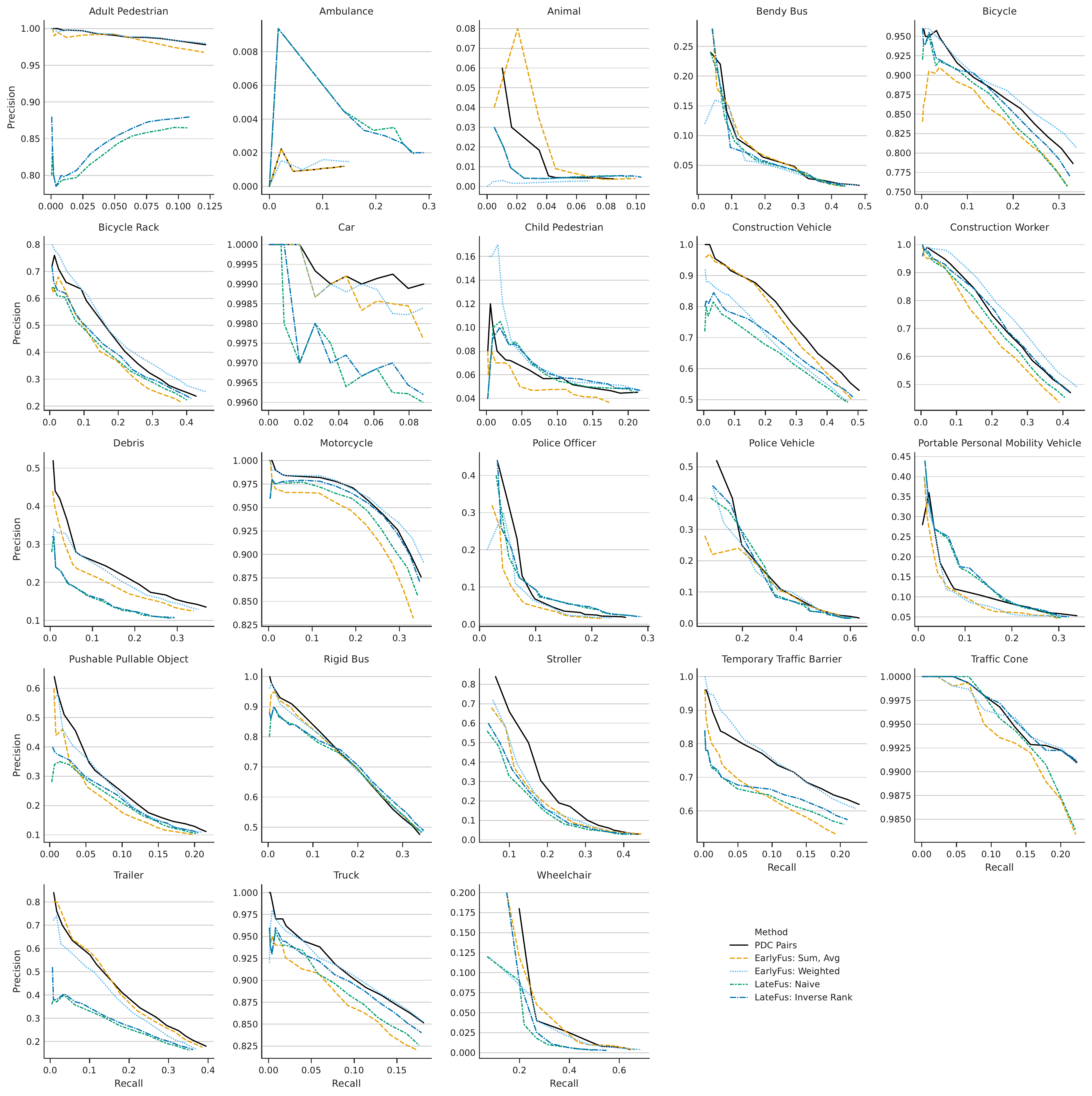}
\label{fig:diagA_sup}
\end{figure*}

\begin{table}[h!]
\caption{Comparisons of instruction pairs results fused by various query policies. Average APs across 5 folds is shown per class. Final PDC setup gets the best mAP and best AP for 13 of 23 classes. The next best fusion is \textit{Early: Weighted}, which achieves the best AP for 7 of 23 classes.}
\vspace{-0.1cm}
\label{tab:diagA} 
\vspace{-0.2cm}
\addtolength{\tabcolsep}{-0.2em}
\resizebox{\linewidth}{!}{ 
\begin{tabular}{lccccc|c}
\toprule
Category      &Samps &Sum:Avg &Early:Wt &Late:Naive & Late:Rank &\textbf{PDC Pairs} \\
\midrule 
car&56517&$8.8^{\pm0.1}$&$8.8^{\pm0.1}$&$8.8^{\pm0.1}$&$8.8^{\pm0.1}$&\bm{$8.8^{\pm0.1}$}\\
adult ped.&40241&$11.8^{\pm0.3}$&\bm{$12.1^{\pm0.2}$}&$9.0^{\pm0.3}$&$9.3^{\pm0.3}$&$12.0^{\pm0.2}$\\
truck&23499&$15.4^{\pm1.3}$&$16.5^{\pm1.2}$&$15.7^{\pm0.7}$&$16.1^{\pm0.6}$&\bm{$16.5^{\pm0.7}$}\\
traffic cone&22194&$22.0^{\pm0.3}$&$22.2^{\pm0.2}$&$22.1^{\pm0.3}$&$22.3^{\pm0.2}$&\bm{$22.2^{\pm0.3}$}\\
traffic barrier&13607&$12.9^{\pm3.2}$&$16.9^{\pm2.7}$&$13.3^{\pm2.6}$&$14.0^{\pm2.7}$&\bm{$16.9^{\pm2.3}$}\\
motorcycle&12523&$31.2^{\pm3.2}$&\bm{$34.3^{\pm1.4}$}&$32.4^{\pm1.6}$&$33.2^{\pm1.3}$&$33.6^{\pm1.4}$\\
bicycle&11883&$27.2^{\pm4.2}$&\bm{$30.1^{\pm2.1}$}&$27.5^{\pm3.6}$&$28.2^{\pm3.3}$&$29.1^{\pm2.8}$\\
rigid bus&7042&$24.8^{\pm1.3}$&\bm{$25.0^{\pm1.6}$}&$24.2^{\pm0.9}$&$24.9^{\pm0.9}$&$24.8^{\pm1.1}$\\
construct. wrkr.&5586&$27.6^{\pm4.3}$&\bm{$34.2^{\pm2.9}$}&$29.4^{\pm3.2}$&$31.4^{\pm3.5}$&$31.4^{\pm4.4}$\\
construct. veh.&5258&$36.5^{\pm2.4}$&$32.6^{\pm6.1}$&$30.7^{\pm2.5}$&$33.1^{\pm2.3}$&\bm{$39.1^{\pm1.0}$}\\
bicycle rack&2771&$17.7^{\pm10.6}$&\bm{$22.0^{\pm8.3}$}&$16.7^{\pm6.7}$&$17.6^{\pm6.9}$&$21.3^{\pm10.8}$\\
push-pull object&2585&$4.8^{\pm2.0}$&$5.7^{\pm1.4}$&$4.6^{\pm1.7}$&$4.9^{\pm1.7}$&\bm{$6.2^{\pm2.1}$}\\
trailer&2286&$16.9^{\pm4.5}$&$15.2^{\pm4.9}$&$10.0^{\pm1.7}$&$10.7^{\pm2.0}$&\bm{$17.3^{\pm4.6}$}\\
debris&1840&$7.2^{\pm1.3}$&$7.7^{\pm1.2}$&$4.6^{\pm0.5}$&$4.8^{\pm0.6}$&\bm{$8.9^{\pm1.6}$}\\
child ped.&1060&$1.1^{\pm0.4}$&\bm{$1.8^{\pm0.3}$}&$1.5^{\pm0.6}$&$1.6^{\pm0.6}$&$1.5^{\pm0.4}$\\
pers. mobi. veh.&790&$3.9^{\pm1.8}$&$4.2^{\pm2.2}$&$5.8^{\pm2.3}$&\bm{$6.0^{\pm2.4}$}&$4.5^{\pm1.3}$\\
police officer&356&$3.7^{\pm3.4}$&$4.3^{\pm4.5}$&$5.2^{\pm3.8}$&$5.4^{\pm3.9}$&\bm{$5.5^{\pm5.2}$}\\
stroller&334&$13.6^{\pm8.4}$&$14.0^{\pm8.7}$&$10.0^{\pm2.3}$&$11.0^{\pm2.6}$&\bm{$16.5^{\pm3.9}$}\\
animal&202&\bm{$1.1^{\pm2.0}$}&$0.0^{\pm0.0}$&$0.3^{\pm0.4}$&$0.3^{\pm0.4}$&$0.8^{\pm1.3}$\\
bendy bus&169&$5.3^{\pm4.0}$&$4.5^{\pm2.5}$&$4.9^{\pm2.5}$&$5.9^{\pm3.0}$&\bm{$6.9^{\pm4.4}$}\\
police vehicle&132&$12.3^{\pm6.1}$&$15.6^{\pm9.8}$&$14.6^{\pm11.2}$&$16.0^{\pm11.0}$&\bm{$16.3^{\pm11.0}$}\\
ambulance&40&$0.1^{\pm0.2}$&$0.1^{\pm0.1}$&\bm{$0.4^{\pm0.4}$}&$0.4^{\pm0.4}$&$0.1^{\pm0.2}$\\
wheelchair&33&$13.5^{\pm7.5}$&$11.5^{\pm6.4}$&$6.7^{\pm5.9}$&$9.2^{\pm8.9}$&\bm{$14.7^{\pm9.2}$}\\
\hline
mAP &- & 13.89			&14.75			&12.97			&13.69			&\textbf{15.44} \\
\bottomrule
\end{tabular}
}
\end{table}
\newpage
\textbf{Query Fusion Policy Ablations}
All results that accompany ~\cref{tab:main_diagA} are shown in  ~\cref{tab:diagA} with class PR curves in ~\cref{fig:diagA_sup}. Importantly, we see that across the majority of categories, our final PDC setup with \textit{Sum:Max} outperforms the other fusion methods.  
 
\begin{figure*}[ht!]
\centering
\caption{NuImages Generated Instruction Pairs Diagnostics results: We display the per class AP curve across 5 folds here in parts. These plots correspond to the calculated AP in Tabs. 3 in the main paper. The solid black line is our PDC Pair. We notice that images from our generated pairs contributes more to our final PDC results than text only. We see this particularly in `pushable pullable object', `bendy bus', `construction vehicle', and `temporary traffic barrier'. In the main paper, we see that using both text and images as pairs outperforms the next best (image only) by 1.88 mAP. }
\includegraphics[width=\linewidth]{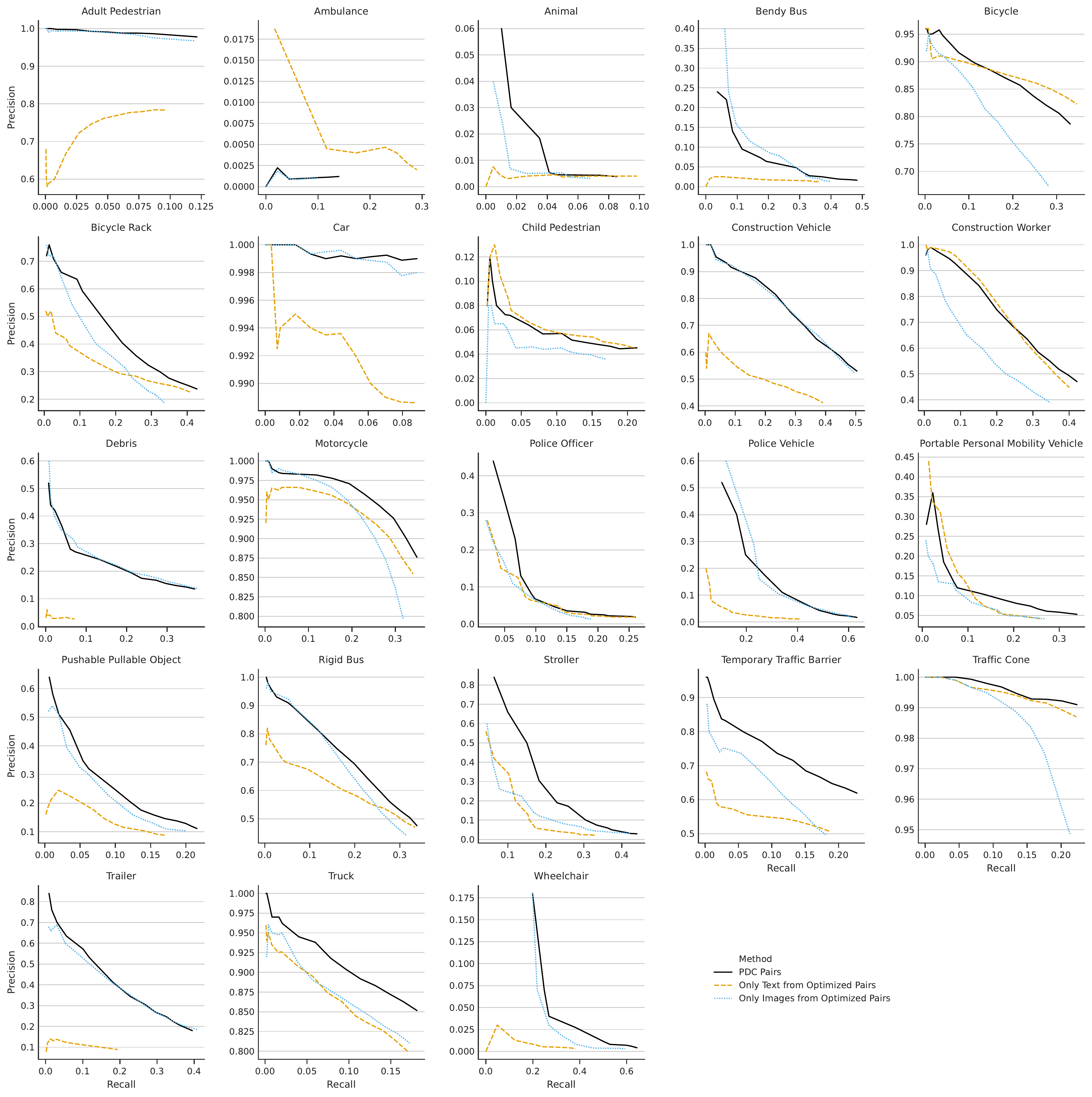}
\label{fig:diagB_sup}
\end{figure*}

\textbf{Generated Instruction Pairs Diagnostics}
All results that accompany ~\cref{tab:main_diagB} are shown in ~\cref{tab:diagB} and class PR curves in ~\cref{fig:diagB_sup}. Again, across most categories, we see that PDC text and bbox pairs outperforms the next best, bbox only. Our main observation holds: using only bboxes is generally better than using only texts. 

\begin{table}[h]
\vspace{-0.2cm}
\caption{Results from using only the texts or bboxes of our PDC instruction set. Average APs across 5 folds is displayed per class. Using both text and bboxes provides the best APs for 17 of 23 classes.}
\label{tab:diagB} 
\vspace{-0.2cm}
\addtolength{\tabcolsep}{-0.2em}
\resizebox{\linewidth}{!}{ 
\begin{tabular}{lccc|c}
\toprule
Category      &Samps &Texts &Bboxes   &\textbf{PDC Pairs} \\
\midrule 
car&56517&$8.7^{\pm0.0}$&$8.8^{\pm0.0}$&\bm{$8.8^{\pm0.1}$}\\
adult pedestrian&40241&$7.2^{\pm0.1}$&$11.8^{\pm0.0}$&\bm{$12.0^{\pm0.2}$}\\
truck&23499&$14.8^{\pm0.2}$&$15.1^{\pm0.1}$&\bm{$16.5^{\pm0.7}$}\\
traffic cone&22194&$22.1^{\pm0.1}$&$21.1^{\pm0.3}$&\bm{$22.2^{\pm0.3}$}\\
temporary traffic barrier&13607&$10.6^{\pm0.6}$&$12.3^{\pm0.7}$&\bm{$16.9^{\pm2.3}$}\\
motorcycle&12523&$32.0^{\pm0.1}$&$30.1^{\pm0.6}$&\bm{$33.6^{\pm1.4}$}\\
bicycle&11883&\bm{$30.4^{\pm0.1}$}&$23.5^{\pm1.3}$&$29.1^{\pm2.8}$\\
rigid bus&7042&$20.6^{\pm0.3}$&$22.8^{\pm0.1}$&\bm{$24.8^{\pm1.1}$}\\
construction worker&5586&$30.5^{\pm0.2}$&$21.4^{\pm1.0}$&\bm{$31.4^{\pm4.4}$}\\
construction vehicle&5258&$20.1^{\pm0.6}$&$38.6^{\pm0.5}$&\bm{$39.1^{\pm1.0}$}\\
bicycle rack&2771&$13.5^{\pm0.3}$&$16.7^{\pm2.2}$&\bm{$21.3^{\pm10.8}$}\\
pushable pullable object&2585&$2.8^{\pm0.2}$&$5.4^{\pm0.4}$&\bm{$6.2^{\pm2.1}$}\\
trailer&2286&$2.4^{\pm0.2}$&$16.7^{\pm0.7}$&\bm{$17.3^{\pm4.6}$}\\
debris&1840&$0.3^{\pm0.0}$&\bm{$9.3^{\pm0.2}$}&$8.9^{\pm1.6}$\\
child pedestrian&1060&\bm{$1.6^{\pm0.1}$}&$0.9^{\pm0.1}$&$1.5^{\pm0.4}$\\
portable personal mobility vehicle&790&$4.4^{\pm0.3}$&$3.4^{\pm0.5}$&\bm{$4.5^{\pm1.3}$}\\
police officer&356&$3.9^{\pm0.7}$&$3.8^{\pm0.6}$&\bm{$5.5^{\pm5.2}$}\\
stroller&334&$8.0^{\pm0.5}$&$9.2^{\pm0.7}$&\bm{$16.5^{\pm3.9}$}\\
animal&202&$0.1^{\pm0.0}$&$0.4^{\pm0.1}$&\bm{$0.8^{\pm1.3}$}\\
bendy bus&169&$0.9^{\pm0.1}$&\bm{$8.6^{\pm0.9}$}&$6.9^{\pm4.4}$\\
police vehicle&132&$4.7^{\pm1.1}$&\bm{$18.9^{\pm1.9}$}&$16.3^{\pm11.0}$\\
ambulance&40&\bm{$0.7^{\pm0.2}$}&$0.1^{\pm0.0}$&$0.1^{\pm0.2}$\\
wheelchair&33&$1.2^{\pm0.2}$&$13.2^{\pm2.3}$&\bm{$14.7^{\pm9.2}$}\\
\hline
mAP &- 	&10.50			&13.56			&\textbf{15.44} \\
\bottomrule
\end{tabular}
}
\vspace{-0.2cm}
\end{table}

\vspace{-0.2cm}
\clearpage

\clearpage
\clearpage
\clearpage

\section{COCO Results}
\label{coco_app}
We demonstrate our Proxy Dataset Curator (PDC) results on the COCO dataset. Our implementation setup for COCO is the same as for NuImages as detailed in Sec. 4.1. In the following section, we show three sets of results: 1) APs per class along with mAP, 2) Qualitative results showing the generated instruction pairs from COCO, 3) Per class PR curves used to generate our average APs across folds.


\vspace{-0.2cm}
\subsection{APs}
\vspace{-0.2cm}
We provide the full COCO AP list accompanying Table 2 in the main paper in  ~\cref{tab:coco_base1,tab:coco_base2}. 
\vspace{-0.1cm}
\begin{table}[h!]
\vspace{-0.2cm}
\caption{COCO results part 1: Comparisons of instruction pairs generated by different methods. Average APs across 5 folds is displayed per class. Classes are sorted based on the number of images. mAP is calculated across  all classes. Note, low performing classes have fewer samples. PDC outperforms the next best baselines by a significant 12.9 mAP and performs the best for 75 of 80 classes. We observe that PDC best performs for top 48 most frequent classes. In cases where PDC underperforms, we note that the difference is usually within 1 AP. Lastly, we observe a pattern seen in NuImages, \textit{Random Pairs} outperforms \textit{Random BBoxes} but underperforms both PDC and \textit{Original Texts}.}
\label{tab:coco_base1} 
\addtolength{\tabcolsep}{-0.3em}
\resizebox{\linewidth}{!}{ 
\begin{tabular}{lcccc|c}
\toprule
Category      &Samps &Org. Ts &Rnd. Bbs. &Rnd. Ps &\textbf{PDC Pairs} \\
\midrule 
person&66808&$7.3^{\pm0.1}$&$6.2^{\pm1.3}$&$6.4^{\pm1.2}$& \bm{$7.5^{\pm0.0}$}\\
chair&13354&$15.1^{\pm1.1}$&$3.2^{\pm2.0}$&$7.8^{\pm4.7}$& \bm{$21.5^{\pm1.0}$}\\
car&12786&$18.3^{\pm0.5}$&$3.3^{\pm4.7}$&$5.2^{\pm5.8}$& \bm{$31.3^{\pm1.6}$}\\
dining table&12338&$12.4^{\pm0.5}$&$7.2^{\pm9.4}$&$9.1^{\pm9.8}$& \bm{$25.6^{\pm0.9}$}\\
cup&9579&$16.6^{\pm0.9}$&$2.1^{\pm1.5}$&$6.1^{\pm3.5}$& \bm{$27.7^{\pm1.9}$}\\
bottle&8880&$20.2^{\pm1.5}$&$0.3^{\pm0.3}$&$0.5^{\pm0.4}$& \bm{$31.0^{\pm2.1}$}\\
bowl&7425&$10.1^{\pm1.2}$&$0.9^{\pm0.5}$&$3.3^{\pm1.5}$& \bm{$27.2^{\pm1.5}$}\\
handbag&7133&$9.6^{\pm0.4}$&$0.7^{\pm0.5}$&$1.9^{\pm1.9}$& \bm{$17.9^{\pm1.0}$}\\
truck&6377&$25.6^{\pm0.9}$&$3.2^{\pm4.4}$&$6.6^{\pm5.9}$& \bm{$31.7^{\pm3.3}$}\\
bench&5805&$20.9^{\pm1.9}$&$0.5^{\pm0.5}$&$1.2^{\pm1.0}$& \bm{$28.2^{\pm2.1}$}\\
backpack&5756&$16.9^{\pm1.2}$&$0.2^{\pm0.1}$&$1.3^{\pm1.3}$& \bm{$21.1^{\pm1.6}$}\\
book&5562&$14.0^{\pm0.7}$&$0.3^{\pm0.2}$&$1.0^{\pm0.8}$& \bm{$40.4^{\pm1.5}$}\\
cell phone&5017&$15.2^{\pm1.4}$&$0.6^{\pm0.2}$&$1.3^{\pm0.7}$& \bm{$25.4^{\pm2.5}$}\\
sink&4865&$50.8^{\pm2.0}$&$3.0^{\pm2.0}$&$11.1^{\pm6.2}$& \bm{$60.5^{\pm2.1}$}\\
clock&4863&$61.6^{\pm1.4}$&$8.7^{\pm17.8}$&$26.4^{\pm16.1}$& \bm{$65.6^{\pm1.1}$}\\
tv&4768&$34.2^{\pm1.6}$&$6.7^{\pm9.4}$&$12.5^{\pm12.0}$& \bm{$48.8^{\pm3.1}$}\\
potted plant&4624&$17.7^{\pm1.0}$&$0.4^{\pm0.3}$&$2.6^{\pm2.1}$& \bm{$33.2^{\pm2.1}$}\\
couch&4618&$43.3^{\pm1.7}$&$7.0^{\pm6.4}$&$21.7^{\pm10.2}$& \bm{$47.5^{\pm1.2}$}\\
dog&4562&$23.6^{\pm0.7}$&$18.5^{\pm15.6}$&$26.6^{\pm16.0}$& \bm{$60.5^{\pm3.8}$}\\
knife&4507&$7.8^{\pm1.4}$&$0.9^{\pm0.9}$&$2.0^{\pm1.7}$& \bm{$23.5^{\pm3.7}$}\\
sports ball&4431&$43.9^{\pm2.4}$&$2.2^{\pm4.7}$&$4.5^{\pm9.5}$& \bm{$46.0^{\pm3.6}$}\\
traffic light&4330&$54.5^{\pm2.2}$&$0.3^{\pm0.2}$&$4.9^{\pm5.0}$& \bm{$58.2^{\pm4.8}$}\\
cat&4298&$43.5^{\pm1.5}$&$62.1^{\pm12.7}$&$66.4^{\pm8.9}$& \bm{$82.1^{\pm3.0}$}\\
umbrella&4142&$45.7^{\pm2.9}$&$3.3^{\pm3.2}$&$19.8^{\pm8.0}$& \bm{$51.9^{\pm3.6}$}\\
bus&4141&$64.4^{\pm2.1}$&$26.0^{\pm25.4}$&$36.9^{\pm29.5}$& \bm{$69.2^{\pm3.4}$}\\
tie&3955&$25.2^{\pm2.2}$&$1.0^{\pm0.5}$&$3.7^{\pm2.9}$& \bm{$35.6^{\pm4.5}$}\\
bed&3831&$52.9^{\pm1.5}$&$9.4^{\pm7.5}$&$25.6^{\pm14.1}$& \bm{$56.0^{\pm3.5}$}\\
train&3745&$60.1^{\pm0.7}$&$34.5^{\pm32.6}$&$40.1^{\pm36.0}$& \bm{$79.1^{\pm3.7}$}\\
vase&3730&$20.5^{\pm2.1}$&$0.6^{\pm0.3}$&$1.1^{\pm0.6}$& \bm{$37.2^{\pm3.8}$}\\
\hline
mAP &- & 42.0			&13.4 	&21.1	& \textbf{54.9} \\
\bottomrule
\end{tabular}
}
\vspace{-0.2cm}
\end{table}

\begin{table}[h]
\vspace{-0.2cm}
\caption{COCO results part 2: Comparisons of instruction pairs generated by different methods. Average APs across 5 folds is displayed per class. Classes are sorted based on the number of images containing them.}
\vspace{-0.2cm}
\label{tab:coco_base2} 
\addtolength{\tabcolsep}{-0.3em}
\resizebox{\linewidth}{!}{ 
\begin{tabular}{lcccc|c}
\toprule
Category      &Samps &Org. Ts &Rnd. Bbs. &Rnd. Ps & \textbf{PDC Pairs} \\
\midrule 
spoon&3682&$8.7^{\pm0.9}$&$1.2^{\pm0.7}$&$2.7^{\pm1.9}$& \bm{$19.0^{\pm1.3}$}\\
motorcycle&3661&$75.6^{\pm0.7}$&$24.2^{\pm26.3}$&$42.7^{\pm26.9}$& \bm{$79.2^{\pm1.6}$}\\
surfboard&3635&$31.9^{\pm1.2}$&$8.6^{\pm10.8}$&$18.7^{\pm19.0}$& \bm{$77.7^{\pm8.5}$}\\
skateboard&3603&$70.5^{\pm1.3}$&$1.6^{\pm1.6}$&$14.0^{\pm10.3}$& \bm{$81.3^{\pm5.3}$}\\
tennis racket&3561&$76.4^{\pm0.8}$&$47.8^{\pm41.7}$&$59.0^{\pm32.2}$& \bm{$93.1^{\pm2.3}$}\\
toilet&3502&$82.9^{\pm1.7}$&$27.7^{\pm32.9}$&$41.8^{\pm33.6}$& \bm{$83.9^{\pm1.6}$}\\
bicycle&3401&$52.5^{\pm2.2}$&$17.6^{\pm18.2}$&$32.5^{\pm14.7}$& \bm{$56.8^{\pm4.4}$}\\
bird&3362&$23.1^{\pm1.1}$&$10.1^{\pm13.1}$&$12.3^{\pm15.2}$& \bm{$53.0^{\pm2.7}$}\\
pizza&3319&$81.2^{\pm0.8}$&$42.1^{\pm34.4}$&$59.1^{\pm23.4}$& \bm{$88.4^{\pm1.3}$}\\
remote&3221&$13.6^{\pm1.2}$&$0.3^{\pm0.2}$&$0.3^{\pm0.2}$& \bm{$23.6^{\pm2.9}$}\\
skis&3202&$64.4^{\pm3.5}$&$4.4^{\pm5.8}$&$18.0^{\pm16.2}$& \bm{$80.2^{\pm1.8}$}\\
boat&3146&$52.0^{\pm2.1}$&$4.8^{\pm6.5}$&$16.9^{\pm11.6}$& \bm{$69.9^{\pm2.7}$}\\
airplane&3083&$40.4^{\pm1.8}$&$49.4^{\pm40.2}$&$52.1^{\pm38.4}$& \bm{$91.9^{\pm2.1}$}\\
horse&3069&$48.8^{\pm1.9}$&$26.9^{\pm8.4}$&$38.7^{\pm5.5}$& \bm{$72.7^{\pm3.4}$}\\
cake&3049&$46.6^{\pm3.9}$&$13.0^{\pm8.7}$&$21.9^{\pm12.1}$& \bm{$69.0^{\pm5.3}$}\\
oven&2992&$46.4^{\pm0.6}$&$9.5^{\pm15.7}$&$15.5^{\pm19.4}$& \bm{$63.8^{\pm3.2}$}\\
baseball glove&2729&$49.2^{\pm1.5}$&$0.2^{\pm0.1}$&$2.4^{\pm1.8}$& \bm{$82.1^{\pm1.8}$}\\
giraffe&2647&\bm{$97.3^{\pm0.4}$}&$70.1^{\pm36.4}$&$86.7^{\pm19.4}$&$96.6^{\pm1.1}$\\
wine glass&2643&$45.0^{\pm2.1}$&$6.8^{\pm13.8}$&$17.3^{\pm13.6}$& \bm{$58.6^{\pm2.6}$}\\
baseball bat&2603&$11.9^{\pm1.7}$&$3.2^{\pm2.3}$&$5.5^{\pm4.1}$& \bm{$69.0^{\pm1.3}$}\\
suitcase&2507&$44.2^{\pm3.1}$&$1.0^{\pm0.6}$&$6.1^{\pm2.8}$& \bm{$53.0^{\pm4.4}$}\\
sandwich&2463&$43.3^{\pm3.3}$&$15.0^{\pm8.3}$&$24.0^{\pm11.9}$& \bm{$56.8^{\pm4.0}$}\\
refrigerator&2461&$46.2^{\pm1.8}$&$2.5^{\pm1.6}$&$9.5^{\pm5.2}$& \bm{$47.9^{\pm4.5}$}\\
kite&2352&$47.5^{\pm1.6}$&$5.2^{\pm6.0}$&$12.4^{\pm13.1}$& \bm{$68.5^{\pm6.7}$}\\
banana&2346&$63.4^{\pm3.3}$&$19.2^{\pm25.1}$&$30.7^{\pm27.7}$& \bm{$71.8^{\pm3.0}$}\\
frisbee&2268& \bm{$36.3^{\pm3.4}$}&$0.4^{\pm0.4}$&$0.8^{\pm1.2}$&$28.1^{\pm4.8}$\\
teddy bear&2234&$64.9^{\pm3.0}$&$19.0^{\pm23.2}$&$45.0^{\pm13.6}$& \bm{$77.1^{\pm2.4}$}\\
elephant&2232&$87.5^{\pm1.8}$&$84.3^{\pm4.3}$&$89.9^{\pm3.2}$& \bm{$91.7^{\pm4.1}$}\\
keyboard&2221&$49.5^{\pm3.1}$&$13.1^{\pm19.7}$&$26.7^{\pm21.9}$& \bm{$57.1^{\pm4.6}$}\\
cow&2055&$56.6^{\pm1.5}$&$29.7^{\pm22.8}$&$42.2^{\pm21.6}$& \bm{$67.9^{\pm5.6}$}\\
broccoli&2010&$70.1^{\pm2.4}$&$35.6^{\pm23.9}$&$56.6^{\pm26.8}$& \bm{$85.0^{\pm1.5}$}\\
zebra&2001& \bm{$98.5^{\pm0.5}$}&$71.5^{\pm36.7}$&$93.1^{\pm8.6}$&$97.9^{\pm0.7}$\\
mouse&1964&$5.1^{\pm1.1}$&$1.0^{\pm1.0}$&$2.3^{\pm3.7}$& \bm{$54.0^{\pm1.5}$}\\
stop sign&1803& \bm{$69.9^{\pm1.7}$}&$16.4^{\pm19.1}$&$41.8^{\pm16.2}$&$68.4^{\pm1.4}$\\
fire hydrant&1797&$69.6^{\pm1.0}$&$0.9^{\pm1.3}$&$12.9^{\pm11.7}$& \bm{$61.8^{\pm2.6}$}\\
orange&1784&$45.9^{\pm1.0}$&$5.0^{\pm5.6}$&$16.2^{\pm10.5}$& \bm{$64.3^{\pm5.1}$}\\
carrot&1764&$56.9^{\pm2.3}$&$2.0^{\pm1.9}$&$10.8^{\pm8.2}$& \bm{$65.5^{\pm2.3}$}\\
snowboard&1703& \bm{$48.2^{\pm5.0}$}&$0.3^{\pm0.1}$&$8.6^{\pm3.4}$&$47.3^{\pm3.3}$\\
apple&1662&$26.6^{\pm1.7}$&$5.8^{\pm6.7}$&$14.2^{\pm11.4}$& \bm{$56.9^{\pm3.3}$}\\
microwave&1601&$40.0^{\pm3.9}$&$1.2^{\pm0.8}$&$5.5^{\pm3.4}$& \bm{$42.3^{\pm2.3}$}\\
sheep&1594&$78.3^{\pm0.7}$&$40.6^{\pm25.2}$&$59.3^{\pm20.4}$& \bm{$85.4^{\pm1.4}$}\\
donut&1585&$56.6^{\pm3.1}$&$7.4^{\pm15.7}$&$11.0^{\pm19.7}$& \bm{$69.7^{\pm3.2}$}\\
hot dog&1273&$49.5^{\pm1.7}$&$20.8^{\pm18.8}$&$28.2^{\pm19.6}$& \bm{$59.2^{\pm7.3}$}\\
toothbrush&1041&$22.0^{\pm2.5}$&$0.3^{\pm0.2}$&$0.6^{\pm0.4}$& \bm{$34.1^{\pm4.5}$}\\
bear&1009&$56.8^{\pm2.3}$&$66.6^{\pm18.6}$&$72.5^{\pm14.0}$& \bm{$85.0^{\pm2.4}$}\\
scissors&975&$23.2^{\pm2.7}$&$0.4^{\pm0.3}$&$2.9^{\pm4.6}$& \bm{$28.8^{\pm3.8}$}\\
parking meter&742&$38.9^{\pm2.8}$&$0.1^{\pm0.1}$&$0.9^{\pm0.6}$& \bm{$42.3^{\pm4.5}$}\\
toaster&225&$4.6^{\pm4.6}$&$0.4^{\pm0.6}$&$1.1^{\pm2.1}$& \bm{$6.6^{\pm2.7}$}\\
hair drier&198&$11.5^{\pm1.0}$&$0.4^{\pm0.5}$&$0.5^{\pm0.5}$& \bm{$7.1^{\pm4.2}$}\\
\hline
mAP &- & 42.0			&13.4 	&21.1	& \textbf{54.9} \\
\bottomrule
\end{tabular}
}
\vspace{-0.2cm}
\end{table}

\vspace{-0.4cm}
\subsection{PDC Generated Instruction Pairs}
\begin{figure*}[h!]
\centering
\caption{COCO results \textit{part 1}: PDC's generated COCO instruction pairs are shown. For simplicity, we display the common text for each pair on the left side instead of over each image. The order in which (text,image) pairs are selected is shown from left to right. Next, we observe that the generated instruction pairs show 1) prototypical images (seen in all categories displayed), 2) corner case images, 3) occluded objects. In this set we see particularly interesting corner cases. For instance, `sandwich' instruction set contains a `sub' that looks remarkably like a `hot dog'. We see various types of `stop signs' - graffitied, in Arabic, and even crumpled. We also see an unique example of a sheep that is currently being sheared, a picture of a sheep, and a flock of sheep. In `potted plant', we observe various different types of plants in various sizes and vases. Lastly, we emphasize that while we only show instruction pairs for 10 categories, all categories show remarkably interesting and visually important details.}
\includegraphics[width=\linewidth]{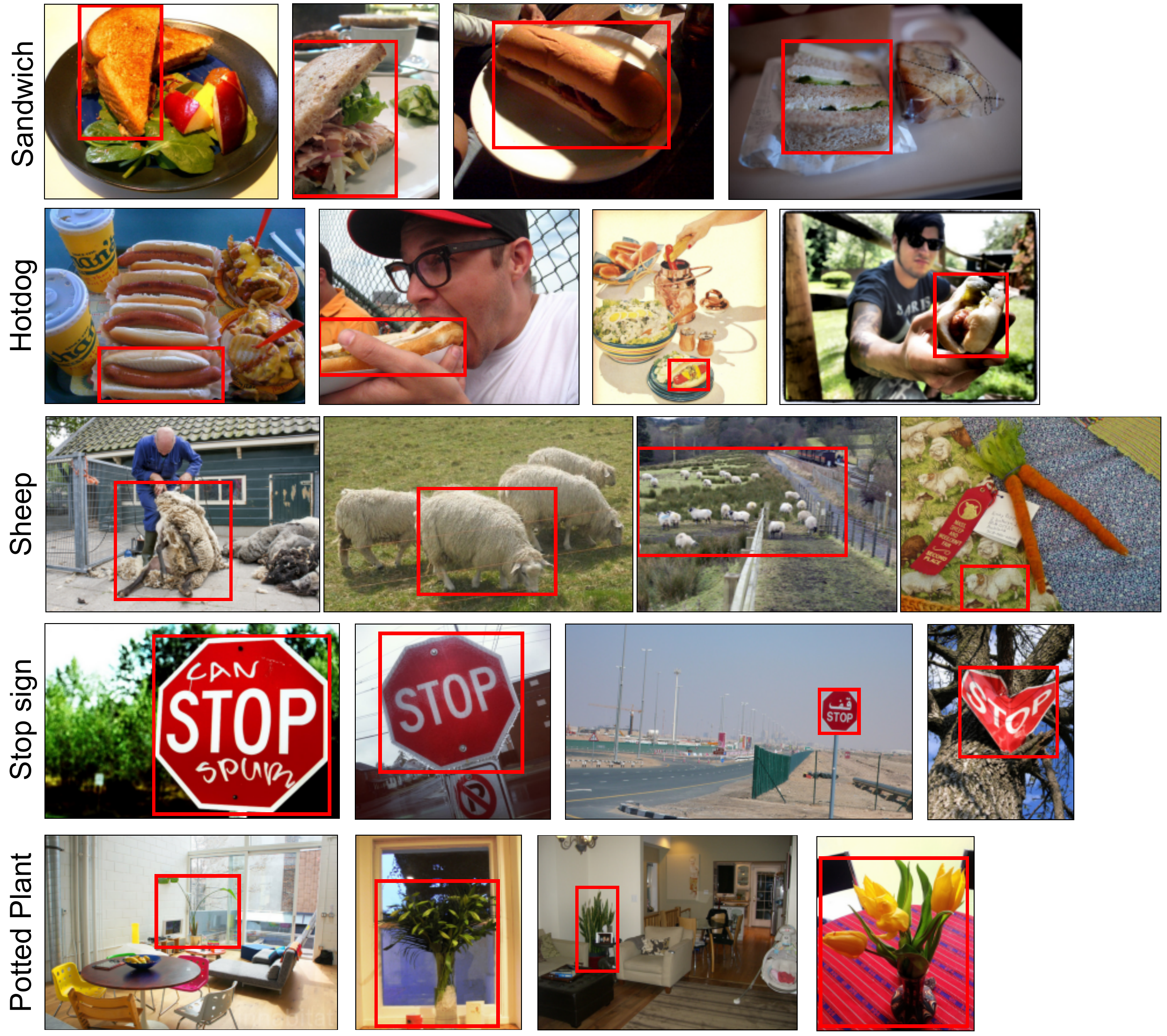}
\label{fig:coco_qual1} 
\vspace{-0.2cm}
\end{figure*}

\begin{figure*}[h!]
\centering
\caption{COCO results \textit{part 2}: In this set, we observe examples for object diversity. We observe that each example in `truck' contains a different type: pickup truck, front of semi-tractor, lorry, and a cross between pickup truck and lorry. Additionally, these examples are partly occluded. Interestingly, we see unique yellow and occluded `fire hydrant'. We also see a rare red and white colored fire hydrant as well as a black and white image. In `sports ball' and `dog', we again observe various types of objects in different viewpoints: volley ball, tennis ball, and soccer ball; back of a dog's head, boxer dog, and dog jumping. Lastly, we note that each example in `backpack' is of different color, size, and viewpoint. }
\includegraphics[width=\linewidth]{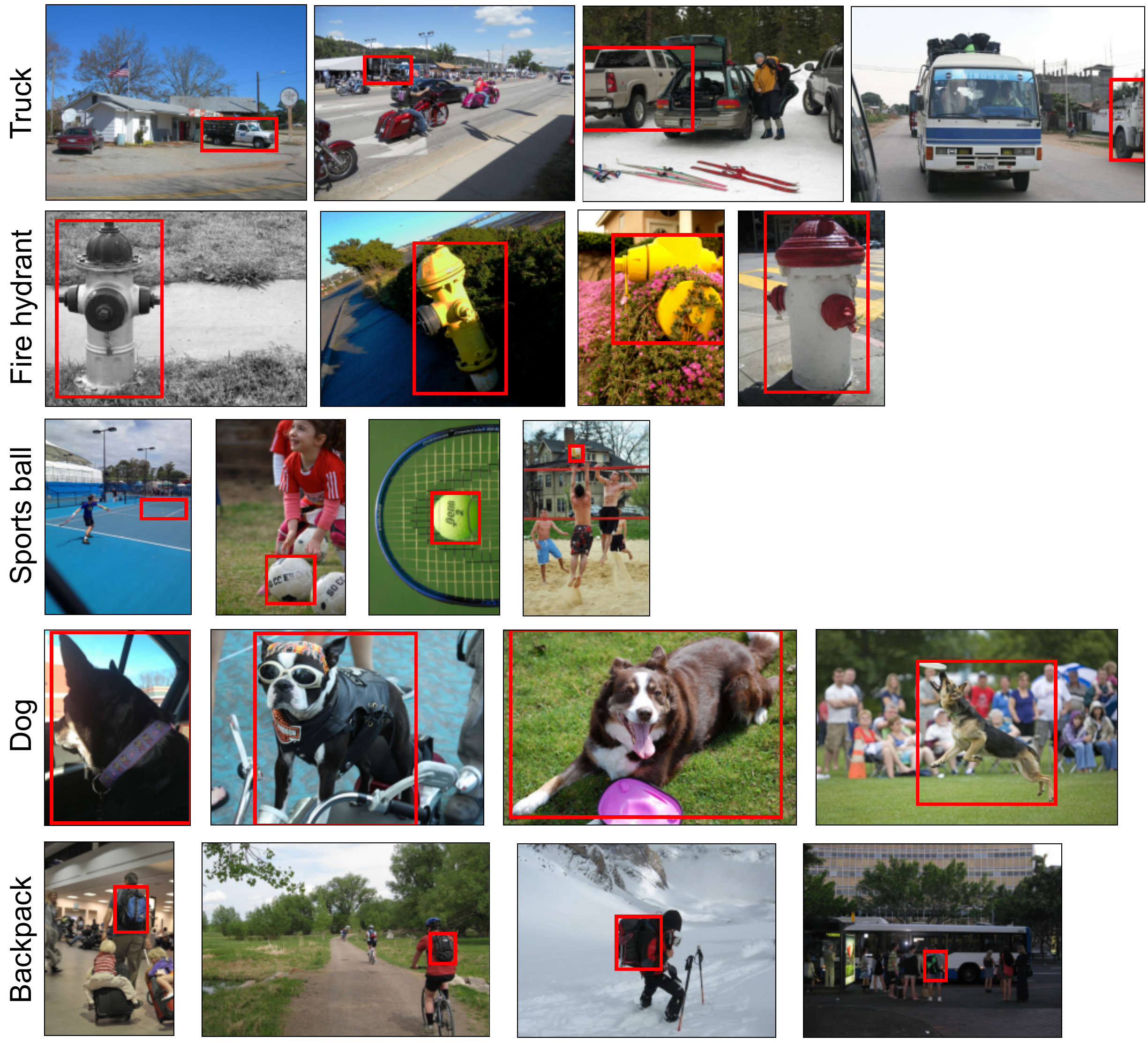}
\label{fig:coco_qual2} 
\vspace{-0.2cm}
\end{figure*}

\vspace{-0.2cm}
We show a subset of our generated instruction pairs for COCO in ~\cref{fig:coco_qual1,fig:coco_qual2} since we have too many qualitative results that can reasonably be displayed in a paper. 

\vspace{-0.2cm}
\subsection{Class PR curves.} 
\begin{figure*}[hbtp]
\centering
\caption{COCO results \textit{part 1}:  We display the per class AP curve across 5 folds here in parts. These plots correspond to the calculated AP in ~\cref{tab:coco_base1,tab:coco_base2}. The solid black line is our PDC Pair. We note that for each class, our PDC curve is comfortably above the others. Furthermore, we see that our precisions and recall are quite high, indicating that we find a significant amount of ground truth objects in the top retrievals. }
\includegraphics[width=\linewidth]{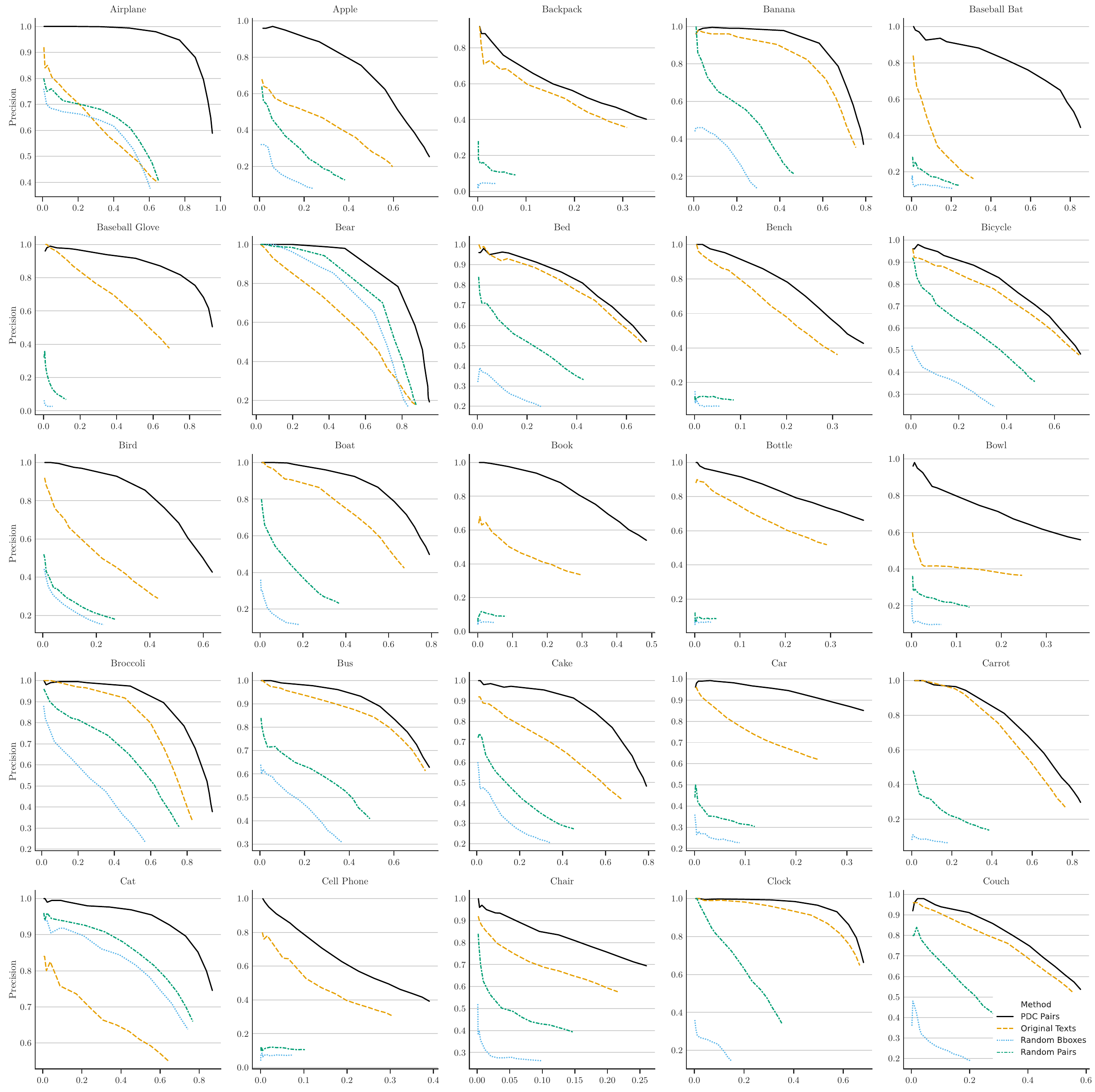}
\label{fig:coco_pr1} 
\vspace{-0.2cm}
\end{figure*}

\begin{figure*}[hbtp]
\centering
\caption{COCO results \textit{part 2}:  We display the per class AP curve across 5 folds here in parts. These plots correspond to the calculated AP in ~\cref{tab:coco_base1,tab:coco_base2}. The solid black line is our PDC Pair. Again, we see that for each class, our PDC curve is comfortably above the others. Precisions and recalls remain high, indicating that we find a significant amount of ground truth objects in the top retrievals.}
\includegraphics[width=\linewidth]{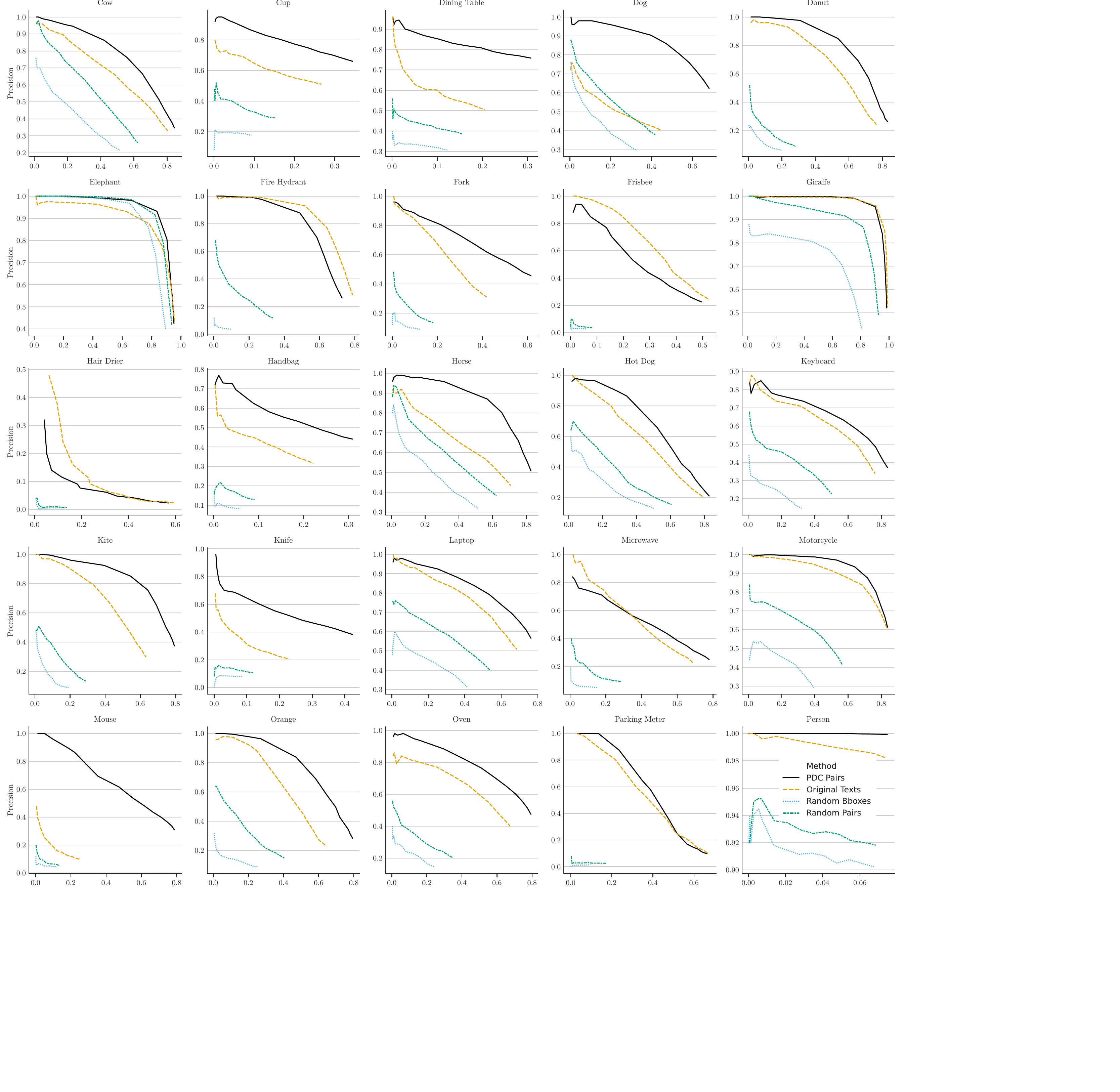}
\label{fig:coco_pr2} 
\vspace{-0.2cm}
\end{figure*}

\begin{figure*}[hbtp]
\centering
\caption{COCO results \textit{part 3}:  We display the per class AP curve across 5 folds here in parts. These plots correspond to the calculated AP in ~\cref{tab:coco_base1,tab:coco_base2}. The solid black line is our PDC Pair. Again, we see that for each class, our PDC curve is comfortably above the others. Precisions and recalls remain high, indicating that we find a significant amount of ground truth objects in the top retrievals.}
\includegraphics[width=\linewidth]{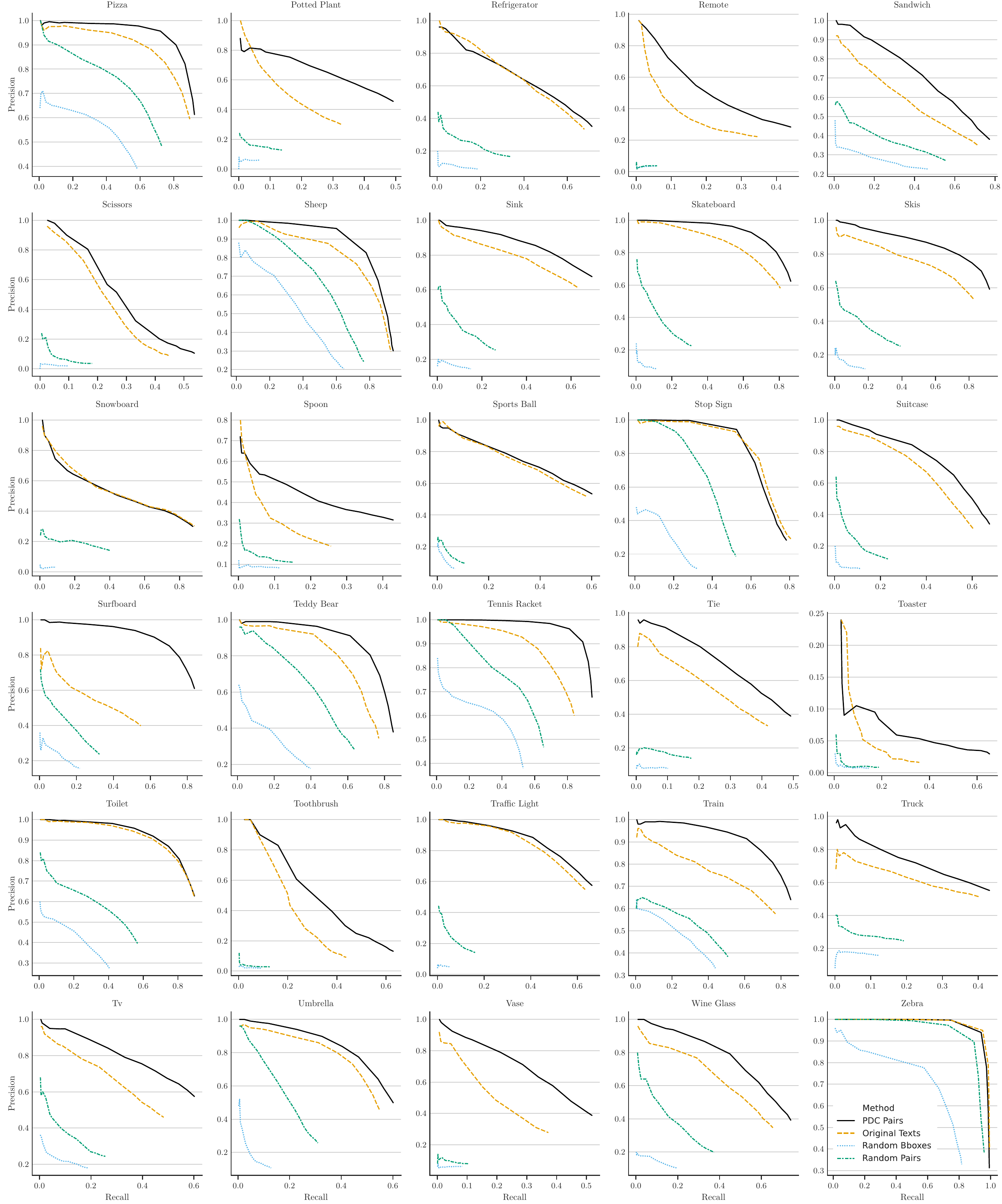}
\label{fig:coco_pr3} 
\vspace{-0.2cm}
\end{figure*}

\vspace{-0.2cm}
We provide class PR curves for all COCO classes in ~\cref{fig:coco_pr1,fig:coco_pr2,fig:coco_pr3}. 

\vspace{-0.2cm}

\clearpage
\vspace{-0.2cm}

\clearpage

\vspace{-0.2cm}

\vspace{-0.2cm}

\vspace{-0.2cm}

\end{document}